%% file: 0.tex
\documentclass{article} % For LaTeX2e
\usepackage{iclr2027_conference,times}
\iclrfinalcopy
\input{math_commands.tex}

\usepackage{hyperref}
\usepackage{url}
\usepackage{amsmath,amssymb}
\usepackage{booktabs}
\usepackage{graphicx}
\usepackage[table]{xcolor}
\usepackage{placeins}
\usepackage{multirow} 
\usepackage{enumitem}
\usepackage{algorithm} 
\usepackage{algorithmicx}
\usepackage{algpseudocode}
\usepackage{eso-pic}
\usepackage{fancyhdr}

\title{On Unlearning for Time-series Forecasting}

\author{\textbf{Zeyu Shi}$^{1,*}$ \quad
\textbf{Yanhui Luo}$^{2,*}$ \quad
\textbf{Ziming Hong}$^{3}$ \quad
\textbf{Chongyang Gao}$^{4,5}$ \\
\textbf{Kezhen Chen}$^{5}$ \quad
\textbf{Shanshan Ye}$^{6}$ \quad
\textbf{Lixu Wang}$^{1,\dagger}$
\\[0.6em]
$^{1}$The Chinese University of Hong Kong, Shenzhen
\quad
$^{2}$Zhejiang University
\\
$^{3}$University of Sydney
\quad
$^{4}$Northwestern University
\quad
$^{5}$Analogy AI, Inc.
\\
$^{6}$Mohamed Bin Zayed University of Artificial Intelligence
\\[0.5em]
{\small
\texttt{\{shizeyu8023,hoongzm,kzchen0204\}@gmail.com}
\quad
\texttt{3230101757@zju.edu.cn}
}
\\
{\small
\texttt{chongyanggao2026@u.northwestern.edu}
\quad
\texttt{shanshan.ye@mbzuai.ac.ae}
}
\\
{\small
\texttt{wanglixu@cuhk.edu.cn}
}
\\[0.4em]
}

\fancypagestyle{preprintpages}{%
    \fancyhf{}
    \fancyhead[L]{\textit{Preprint. Under review.}}
    \fancyfoot[C]{\thepage}
    \renewcommand{\headrulewidth}{0.4pt}
    \renewcommand{\footrulewidth}{0pt}
}
\fancypagestyle{firstpage}{%
    \fancyhf{}
    \fancyhead[L]{%
        \smash{%
            \raisebox{-0.42cm}{%
                \includegraphics[height=1.5cm]{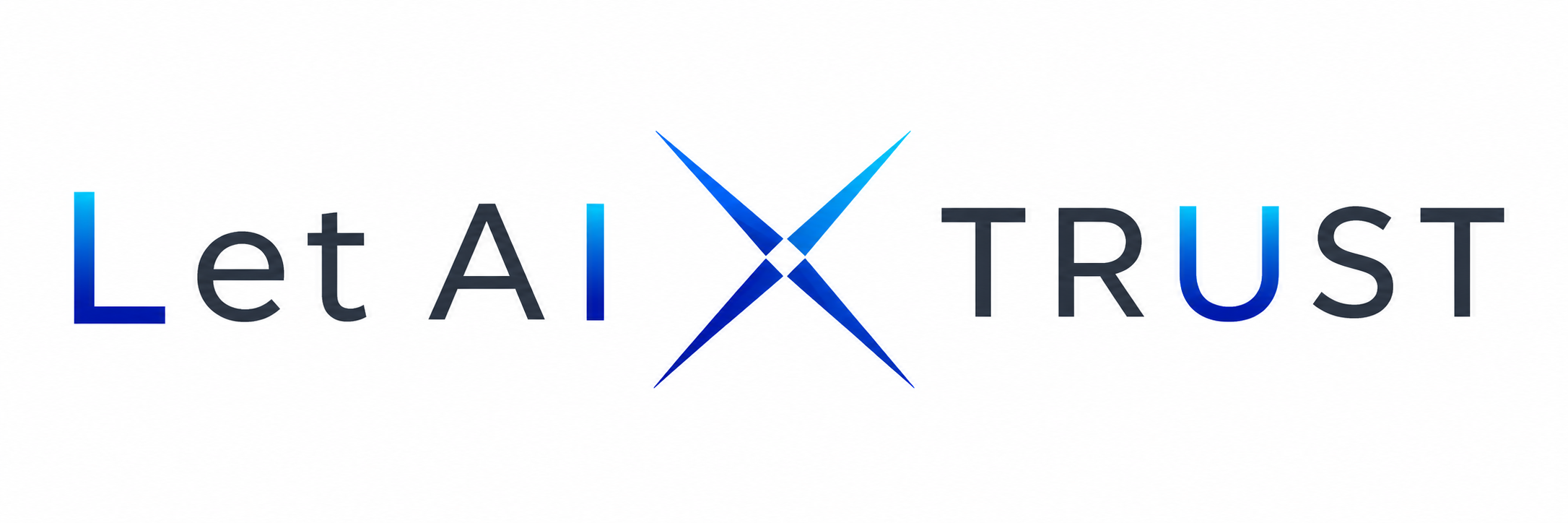}%
            }%
        }%
    }
    \fancyfoot[C]{\thepage}
    \renewcommand{\headrulewidth}{0.4pt}
    \renewcommand{\footrulewidth}{0pt}
}
\begin{document}

\maketitle

\pagestyle{preprintpages}
\thispagestyle{firstpage}

\begingroup
\renewcommand{\thefootnote}{\fnsymbol{footnote}}
\footnotetext[1]{Equal contribution.}
\footnotetext[2]{Corresponding author.}
\endgroup

\begin{abstract}
Time-series forecasting is widely used in healthy, transportation, finance, and many other domains. Models in these settings are often trained on longitudinal user or entity level records, which may later require removal because they contain sensitive or proprietary information or have been corrupted by sensor failures. To address such deletion requests without costly retraining, machine unlearning has been widely studied as a practical mechanism for privacy protection and data governance. However, the application of machine unlearning to time series prediction has not yet been well realized, it mainly dues to the following unique challenges: Gradient-based unlearning can be unstable because a deleted observation participates in multiple causally connected forecasting windows, causing parameter updates to propagate beyond the requested interval and degrade retained forecasting utility. 
Label-guided updating offers a more controlled alternative, but continuous and context-dependent forecasts lack a suitable replacement target, while the exact-retrained output is unavailable during unlearning. Moreover, the remaining support for a deleted temporal pattern is highly non-uniform. Some affected windows retain structurally similar counterparts in the retained data, whereas others
become underrepresented or isolated. We present 
\textbf{\textsc{RDTU}}, a
\underline{\textbf{R}}esidual
\underline{\textbf{D}}iffusion framework of \underline{\textbf{T}}ime-series \underline{\textbf{U}}nlearning that constructs labels without accessing an exact-retrained model at unlearning time. \textsc{RDTU} first uses a retained-set neural tangent kernel predictor to obtain a deletion-compatible base forecast. Then it quantifies the global and local structural support of each affected window using the volume contribution of the retained-reference data.
Conditioned on the query window, retained-set prediction, and structural support, a diffusion model generates a residual correction that estimates the counterfactual forecast, yielding a pseudo-label field that guides a lightweight model update. Experiments across real-world forecasting benchmarks show that \textsc{RDTU} consistently produces unlearned models that most closely match exact retraining among the evaluated efficient unlearning methods. It achieves this alignment while maintaining retained forecasting utility and reducing membership-inference signals.

\end{abstract}

\section{Introduction}
\input{1_intro}

% \section{Related Works}
% \input{2_related}

\section{Problem Definition and Related Works}
\input{3_problem}

\section{Method}
\input{4_method}

\section{Experiments}
\input{5_exp}

\section{Conclusion}
In this work, we introduced RDTU, a label-guided framework for approximate unlearning in time-series forecasting. RDTU provides an efficient prediction by constructing an NTK predictor, and the structural irrelevance score is used to characterize the global and local support conditions of each affected window. Conditional residual diffusion subsequently corrects the discrepancy between the NTK approximation and the corresponding neural predictor, and the resulting labels guide a lightweight unlearning update.  Experiments on four univariate and multivariate forecasting benchmarks show that RDTU achieves the lowest F-P Gap and test NRMSE among the evaluated efficient unlearning methods, while maintaining low membership distinguishability and requiring substantially less computation than exact retraining.

\clearpage

\section*{AI use statement}
Generative AI tools were used solely to support the linguistic editing of this manuscript, including grammar correction, sentence rephrasing, readability improvement, and refinement of textual flow. They were not used to formulate research questions or hypotheses, develop the methodology, design or conduct experiments, analyze data, interpret results, or generate scientific conclusions. All AI-assisted text was critically reviewed and revised by the authors, who take full responsibility for the accuracy, originality, and integrity of the final manuscript.

\section*{Ethics statement}
This work investigates machine unlearning as a means of reducing the influence of designated observations on time-series forecasting models. All experiments use existing publicly available benchmark datasets; no new participants were recruited, and no new personal data were collected. The EEG and HAR datasets contain previously collected human-subject measurements, and this study uses only the publicly released research data. RDTU provides approximate rather than certified unlearning and should not be interpreted as offering a formal or legal guarantee of data erasure. Its effectiveness depends on the adopted threat model, evaluation protocol, and deployment setting. Applications in privacy-sensitive domains should therefore incorporate appropriate access controls, deletion audits, and regulatory review. Moreover, deletion requests and model updates should be documented and independently auditable to prevent unlearning techniques from being misused to obscure data provenance or selectively manipulate model behaviour.

\section*{Reproducibility statement}
To support reproducibility, all methods were tested using the same preprocessing, data partitioning, deletion requests, and evaluation procedures. The detailed experimental configurations and implementation methods are listed in the appendix. For each dataset, we evaluated 10 sequences or subjects under three deletion seeds, averaged the results for each sequence or subject under different seeds, and then reported the corresponding macro average. The source code, configuration files, selected sequence or subject identifiers, and random seeds will be included in the supplementary materials. The published content will include scripts for training the original model and re-training the precise model, as well as scripts for executing our experiments.

\clearpage
\bibliography{iclr2027_conference}
\bibliographystyle{iclr2027_conference}

\clearpage
\appendix
\begin{center}
    {\Large\bfseries Appendix}
\end{center}
\input{6_Appendix}
% Required packages:
% \usepackage{booktabs}
% \usepackage{multirow}
% \usepackage{graphicx}

% Required packages:
% \usepackage{booktabs}
% \usepackage{multirow}
% \usepackage{graphicx}

\end{document}

%% file: math_commands.tex
\usepackage{amsmath,amsfonts,bm}

\def\eqref#1{equation~\ref{#1}}
\def\1{\bm{1}}
\newcommand{\train}{\mathcal{D}}

\def\rvepsilon{{\mathbf{\epsilon}}}

\def\rvr{{\mathbf{r}}}

\def\rvz{{\mathbf{z}}}

\def\vzero{{\bm{0}}}
\def\vone{{\bm{1}}}

\def\vtheta{{\bm{\theta}}}

\def\vc{{\bm{c}}}

\def\vf{{\bm{f}}}
\def\vg{{\bm{g}}}

\def\vk{{\bm{k}}}

\def\vr{{\bm{r}}}

\def\vx{{\bm{x}}}
\def\vy{{\bm{y}}}
\def\vz{{\bm{z}}}

\def\mG{{\bm{G}}}

\def\mI{{\bm{I}}}
\def\mJ{{\bm{J}}}
\def\mK{{\bm{K}}}

\def\mR{{\bm{R}}}

\def\mPhi{{\bm{\Phi}}}

\DeclareMathAlphabet{\mathsfit}{\encodingdefault}{\sfdefault}{m}{sl}
\SetMathAlphabet{\mathsfit}{bold}{\encodingdefault}{\sfdefault}{bx}{n}

\def\sI{{\mathbb{I}}}
\def\sJ{{\mathbb{J}}}

\def\sM{{\mathbb{M}}}

\def\sQ{{\mathbb{Q}}}

\def\sS{{\mathbb{S}}}
\def\sT{{\mathbb{T}}}

\newcommand{\E}{\mathbb{E}}
\newcommand{\Ls}{\mathcal{L}}
\newcommand{\R}{\mathbb{R}}

\DeclareMathOperator*{\argmin}{arg\,min}

%% file: 1_intro.tex
Time-series forecasting has become a fundamental component of data-driven decision making across finance, healthcare, transportation, energy systems, and intelligent manufacturing, where historical observations are used to anticipate future risks, demands, and system states~\citep{10.1098/rsta.2020.0209,zonta2020predictive,Mohapatra2024PhasedrivenGR}. As forecasting models are trained on increasingly large and continuously collected data streams, they also face growing demands for selective data removal. In privacy-sensitive domains such as finance and healthcare, training sequences may contain sensitive personal information; under Article~17 of the European Union's General Data Protection Regulation (GDPR), data subjects may request the erasure of their personal data under specified conditions~\citep{eu2016general}. A parallel need arises in industrial monitoring, where sensor failures, calibration drift, and communication faults can introduce erroneous observations, while equipment aging can render previously collected operating patterns obsolete~\citep{teh2020sensor,goknil2023systematic,wang2024dacr,zhao2025exploring}. Once incorporated into model training, such observations may distort the learned temporal dependencies and compromise subsequent forecasts.

To meet these deletion requirements, machine unlearning aims to remove the influence of designated training data from a trained model~\citep{cao2015towards,wang2022nontransferable}. Existing approaches are broadly divided into exact and approximate unlearning. Exact unlearning seeks to match the distribution of models retrained from scratch on the retained data, but often incurs prohibitive computational costs~\citep{NEURIPS2019_cb79f8fa}. Approximate unlearning instead updates the original model to efficiently emulate this behavior. Gradient-ascent methods increase the forgetting-set loss, but their lack of an explicit convergence target can cause unstable optimization and severe utility degradation. NegGrad+~\citep{NEURIPS2023_062d711f} and NPO~\citep{zhang2024negative} mitigate this problem by balancing gradients or bounding the forgetting objective. However, they primarily control how the model departs from forgotten samples rather than specifying its desired post-deletion behavior. This motivates label-based methods such as random labeling, which replace forgotten targets with constructed alternatives~\citep{graves2021amnesiac,di2026label}.

Methods developed for image and text unlearning do not transfer directly to time-series forecasting. Forecasting windows are ordered and extensively overlapping, so gradient ascent on forgotten windows may disrupt temporal representations shared with retained data. Label-based methods also lack a natural replacement target because arbitrary continuous values may violate temporal and cross-variable structures. Unlike classification labels or abstention responses~\citep{graves2021amnesiac,maini2024tofu}, time series have no universal rejection label. TS-Unlearn addresses this problem by feeding Gaussian inputs into the original model to construct forgetting targets and dynamically balancing forgetting and retention~\citep{wang2026ts}. However, these targets are not conditioned on the specific query or its local support in the retained data. They constrain outputs to a plausible range but cannot approximate the predictions of exact unlearning.

To address these challenges, we adopt label-based approximate unlearning and construct labels that approximate the predictions of exact unlearning for time-series forecasting. Because obtaining such labels directly would require retraining on the strictly retained set, we use a lightweight neural tangent kernel (NTK) predictor as an efficient regression surrogate. However, finite-width networks perform feature learning and nonlinear optimization, creating a non-negligible gap between their predictions and the NTK approximation. We therefore introduce training a small conditional diffusion model to learn the residual distribution between the trained model and its NTK predictor and transfer this residual pattern to retained-set conditions. To avoid conditioning on high-dimensional raw time-series windows, we further propose the structural irrelevance score (SIS), which summarizes complementary global directional change and local retained support. Conditioned on the retained-set NTK prediction and SIS, diffusion generates a residual correction that forms a suitable label for each affected window. Extensive experiments on univariate and multivariate datasets demonstrate a favorable balance among unlearning effectiveness, forecasting utility, and computational efficiency. Our contributions can be summarized as follows:
\begin{itemize}[leftmargin=*]
\item We formulate time-series forecasting unlearning under strict temporal deletion closure, which identifies all forecasting windows affected by a deletion request, and cast efficient approximate unlearning as the construction of counterfactual labels that approximate the predictions of an exact-retrained model.

\item We introduce the Structural Irreplaceability Score (SIS), which complements the model-dependent NTK prediction by quantifying the global and local structural support of each affected window. SIS provides a compact condition for residual generation and reduces reliance on costly raw-window representations.

\item We propose \textbf{RDTU}, a label-guided Residual Diffusion framework that combines a retained-set NTK predictor with a simple conditional residual diffusion model. The NTK predictor provides a lightweight counterfactual base prediction, while the diffusion model compensates for the substantial finite-width discrepancy between the NTK approximation and the corresponding trained neural predictor.
\end{itemize}

%% file: 3_problem.tex
\subsection{Sliding-Window Forecasting}
\label{subsec:forecasting}

Let \(\{x_1,\ldots,x_i,\ldots,x_T\}\) denote the observed time series to train a forecasting model \footnote{We focus on univariate data to simplify notation, and the concepts can be extended to multivariate ones.}. Given a lookback length \(\ell\) and a forecasting horizon \(h\), each valid forecast origin \(t\) induces a supervised forecasting sample \(\vx_t= (x_{t-\ell},...,x_{t-1})\) and
\(\vy_t=(x_t,...,x_{t+h-1})\). The original training set is \(\train=\{(\vx_t,\vy_t)\}_{t\in\sT_{\mathrm{train}}}\),
where \(\sT_{\mathrm{train}}\subseteq\{\ell+1,\ldots,T-h+1\}\) is the set of valid training forecast origins. An original forecasting model is trained as
\begin{equation}
\vtheta_{\train}^{*}
=
\argmin_{\vtheta}
\mathcal{L}
\left(
\vtheta,
\train
\right),
\qquad
\widehat{\vy}_t
=
f_{\vtheta_{\train}^{*}}
\left(
\vx_t
\right).
\label{eq:original-model}
\end{equation}
where $\mathcal{L}(\vtheta,\train)$ denotes the empirical forecasting loss, which we usually instantiate as the mean squared error between the predicted and ground-truth future windows~\citep{NEURIPS2021_bcc0d400}.

\subsection{Time-series Exact Unlearning and Approximate Unlearning}
\label{subsec:strict-unlearning}

Consider an unlearning request for a contiguous temporal interval
\(\sT_{\mathrm{f}}=[t_{\mathrm{i}},t_{\mathrm{s}})\) in the original time series. The request asks the model to forget the training influence of the observations
${x_{t_{\mathrm{i}}},x_{t_{\mathrm{i}}+1},\ldots,x_{t_{\mathrm{s}}-1}}$. For time-series forecasting, this unlearning request cannot be handled by simply
removing the training windows whose prediction targets lie inside $\sT_{\mathrm{f}}$. Each supervised window contains both an input context and a future prediction
target. Therefore, an observation in $\sT_{\mathrm{f}}$ may appear as the target of one training window, while also appearing as part of the input context of several
subsequent windows.

For a forecasting sample \((\vx_t,\vy_t)\), its complete
temporal footprint is \(\sI_t=[t-\ell,t+h)\),
which includes its input interval $[t-\ell,t)$ and target interval $[t,t+h)$. We regard a sample as affected by the deletion request whenever any part of its temporal footprint overlaps the requested interval:
\begin{equation}
\train_{\mathrm{f}}(\sT_{\mathrm{f}})
=
\left\{
(\vx_t,\vy_t)\in\train
\;\middle|\;
\sI_t\cap \sT_{\mathrm{f}}\neq\varnothing
\right\}.
\label{eq:strict-closure}
\end{equation}
We call $\train_{\mathrm{f}}(\sT_{\mathrm{f}})$ the \emph{strict temporal deletion closure} of $\sT_{\mathrm{f}}$, \emph{i.e.}, unlearning set. Accordingly, the retained training set is not obtained by deleting only the raw observations in $\sT_{\mathrm{f}}$. Instead, it removes every supervised forecasting window affected by those observations: $\train_{\mathrm{r}}(\sT_{\mathrm{f}})=\train\setminus\train_{\mathrm{f}}(\sT_{\mathrm{f}}).$
% \begin{equation}

% \label{eq}
% \end{equation}
% For example, under one-step forecasting with \(\ell=4\), suppose that \(\vx_5\)
% belongs to the requested deletion interval. Then both
% \((\vx_1,\vx_2,\vx_3,\vx_4)\rightarrow \vx_5\) and
% \((\vx_2,\vx_3,\vx_4,\vx_5)\rightarrow \vx_6\) must be removed from the retained
% supervised set: \(\vx_5\) is the target of the first window and an input value
% of the second window.

% \subsection{Deletion-Induced Prediction Trajectory}
% \label{subsec:trajectory}

% Unlike sample-wise unlearning, deleting a raw temporal block induces a
% sequence of coupled prediction queries. Consider one-step forecasting with
% $\ell=4$. The windows
% \begin{equation}
% (x_1,x_2,x_3,x_4)\rightarrow x_5
% \end{equation}
% and
% \begin{equation}
% (x_2,x_3,x_4,x_5)\rightarrow x_6
% \end{equation}
% are temporally coupled because $x_5$ is the target of the first window and an
% input element of the second window. If $x_5$ belongs to a deleted raw block,
% both supervised windows belong to the strict deletion closure.

% Consequently, temporal unlearning cannot be formulated as modifying a single
% label or an isolated training sample. It must specify a coherent set of
% counterfactual outputs over all closure windows. We refer to this set as a
% \emph{pseudo-label field}, which assigns a pseudo-target
% $\tilde{\vy}_t$ to every closure query
% $(\vx_t,\vy_t)\in\train_{\mathrm{f}}$.

% \subsection{Temporal Unlearning Objective}
% \label{subsec:unlearning-objective}
\smallskip \textbf{Exact Unlearning.}
After identifying all windows affected by the deletion request through strict temporal deletion closure, exact unlearning aims to produce a model whose distribution is indistinguishable from that of a model trained without ever observing the forgotten windows~\citep{pmlr-v119-guo20c}. The canonical realization of this objective is exact retraining. Specifically, it needs to instantiate a new forecasting model with the same architecture as the original model, randomly initialize its parameters from the original initialization distribution, and train it from scratch using only the strictly retained set~\citep{NEURIPS2019_cb79f8fa}.

Let $\vtheta_{\train_{\mathrm r},0}$ denote the newly initialized parameters, and let $\rvz_{\train_{\mathrm r}}$ collect the remaining training randomness,
such as the mini-batch order and stochastic regularization. The exact-retrained model and its prediction are defined as
\begin{equation}
\vtheta_{\train_{\mathrm r},0}
\sim
p_{\mathrm{init}},
\qquad
\rvz_{\train_{\mathrm r}}
\sim
p_{\mathrm{rng}},
\qquad
\vtheta_{\train_{\mathrm r}}^{*}
=
\mathcal{A}
\left(
\train_{\mathrm r};
\vtheta_{\train_{\mathrm r},0},
\rvz_{\train_{\mathrm r}}
\right).
\label{eq:exact-retrain-parameters}
\end{equation}

where $\mathcal{A}$ denotes the original training algorithm, $p_{\mathrm{init}}$ denotes the parameter-initialization distribution, and $p_{\mathrm{rng}}$ denotes the joint distribution of the remaining training
randomness, including mini-batch sampling and stochastic regularization.
% \begin{equation}
% \vy_t^{*}
% =
% f_{\vtheta_{\mathrm r}^{*}}
% \left(
% \vx_t
% \right).
% \label{eq:exact-retrain-prediction}
% \end{equation}

% An unlearning algorithm is considered exact when applying it to the original model produces the same model distribution as exact retraining on $\train_{\mathrm{r}}$. Formally, let the original model be $\vtheta^{}=A(\train;\xi)$, where
% $\xi \sim P_{\Xi}$. An unlearning algorithm $U$ satisfies exact unlearning if
% \begin{equation}
% U
% \left(
% \vtheta^{},
% \train,
% \sT_{\mathrm f}
% \right)
% \overset{d}{=}
% A
% \left(
% \train_{\mathrm r};
% \xi_{\mathrm r}
% \right),
% \label{eq}
% \end{equation}
% where $(\overset{d}{=})$ denotes equality in distribution over the randomness of training and unlearning.
% \lixu{The introduction of exact unlearning is not enough, we should introduce the process of randomly initializing another model with the same architecture. Refer to the standard definition of exact unlearning in some machine unlearning survey papers.}

\smallskip \textbf{Approximate Unlearning.}
Conducting a complete model retraining or training an exact unlearning model might be inefficient and costly in many cases. Given the original model
\(f_{\vtheta^*_\train}\) and a deletion request \(\sT_{\mathrm{f}}\), an unlearning algorithm
produces \(\vtheta_{\mathrm{u}}=\mathcal{U}(\vtheta^*,\train,\sT_{\mathrm{f}})\) and aims to make its forecasting behavior close to that of Exact Retrain:
\begin{equation}
\E_{\vx\sim P_{\sQ}}
\left[
\left\|
f_{\vtheta_{\mathrm{u}}}(\vx)
-
f_{\vtheta_{\mathrm{r}}^{*}}(\vx)
\right\|_{\mathrm{F}}^2
\right]
\approx 0.
\label{eq:functional-equivalence}
\end{equation}
where $P_{\sQ}$ denotes a distribution of permissible forecasting queries.

\smallskip
\noindent\textbf{Gradient-Ascent-Based Approximate Unlearning.}
Gradient-ascent-based methods increase the training loss on the forgetting set $\train_{\mathrm f}$, thereby weakening the associations learned from the
samples to be removed~\citep{gao2025on}. To preserve the remaining knowledge, gradient ascent on
the forgetting set is commonly combined with gradient descent on the retained set $\train_{\mathrm r}$. The resulting optimization objective can be written as
\begin{equation}
\vtheta_{\mathrm{u},\mathrm{GA}}^{(0)}
=
\vtheta_{\train}^{*},
\qquad
\vtheta_{\mathrm{u},\mathrm{GA}}
=
\argmin_{\vtheta}
\left[
\mathcal{L}
\left(
\vtheta,
\train_{\mathrm r}
\right)
-
\lambda_{\mathrm f}
\mathcal{L}
\left(
\vtheta,
\train_{\mathrm f}
\right)
\right].
\label{eq:gradient-ascent-unlearning}
\end{equation}
where $\lambda_{\mathrm f}$ controls the strength of forgetting. The negative forgetting-set term reverses the ordinary learning direction and therefore increases the prediction loss on the affected samples. Although this strategy is computationally efficient, directly maximizing the forgetting loss can make the optimization unstable. In particular, aggressive or prolonged gradient ascent may push the model parameters far away from the original trained solution, causing catastrophic degradation on the retained set or even model collapse~\citep{zhang2024negative}.

\smallskip
\noindent\textbf{Random-Labeling-Based Approximate Unlearning.}
Random-labeling-based methods replace the original targets of the forgetting samples with randomly generated incorrect targets~\citep{graves2021amnesiac}. Let $\vy_t^{\mathrm{rand}}$ denote a random target assigned to an affected window $\vx_t$. The original model is then fine-tuned using the retained samples and
the randomly relabeled forgetting samples:
\begin{equation}
\vtheta_{\mathrm{u},\mathrm{RL}}^{(0)}
=
\vtheta_{\train}^{*},
\qquad
\vtheta_{\mathrm{u},\mathrm{RL}}
=
\argmin_{\vtheta}
\left[
\mathcal{L}
\left(
\vtheta,
\train_{\mathrm r}
\right)
+
\frac{
\lambda_{\mathrm f}
}{
\left|
\train_{\mathrm f}
\right|
}
\sum_{
\left(
\vx_t,
\vy_t
\right)
\in
\train_{\mathrm f}
}
\left\|
f_{\vtheta}
\left(
\vx_t
\right)
-
\vy_t^{\mathrm{rand}}
\right\|_{\mathrm F}^{2}
\right].
\label{eq:random-labeling-unlearning}
\end{equation}

By associating the affected inputs with unrelated targets, random labeling weakens the original input--target relationship without retraining the model from scratch.

\subsection{Neural Tangent Kernel}

\label{subsec:ntk-definition}

The neural tangent kernel (NTK) describes the local training dynamics of a neural network through the gradients of its outputs with respect to its
parameters~\citep{NEURIPS2018_5a4be1fa}. Let
$f_{\vtheta}:\mathbb{R}^{l}\rightarrow\mathbb{R}^{h}$
denote a forecasting model with parameters
$\vtheta\in\mathbb{R}^{d}$, where $l$ is the input-window length, $h$ is the forecasting horizon. At an initialization $\vtheta_0$, the output Jacobian is defined as
\begin{equation}
\mJ_{\vtheta_0}(\vx)
=
\left.
\frac{\partial f_{\vtheta}(\vx)}
{\partial \vtheta^{\top}}
\right|_{\vtheta=\vtheta_0}
\in \mathbb{R}^{h\times d}.
\label{eq:ntk-jacobian}
\end{equation}
The empirical NTK between two input windows $\vx$ and $\vx'$ is the
output-space kernel
\begin{equation}
\mK_{\vtheta_0}(\vx,\vx')
=
\mJ_{\vtheta_0}(\vx)
\mJ_{\vtheta_0}(\vx')^{\top}
\in \mathbb{R}^{h\times h}.
\label{eq:ntk-definition}
\end{equation}
Given a supervised dataset $\sS=\{(\vx_i,\vy_i)\}$, let
$\mK_{\sS\sS}\in\mathbb{R}^{h\times h}$ be the block Gram matrix formed by $\mK_{\vtheta_0}(\vx_i,\vx_j)$, let $\mathbf{k}_{\sS}(\vx)\in\mathbb{R}^{h\times h}$
denote the corresponding cross-kernel matrix, and let
$\vy_{\sS}\in\mathbb{R}^{h}$ contain the stacked vectorized targets. The regularized NTK predictor is defined as
\begin{equation}
g_{\sS}(\vx)
=
\mathbf{k}_{\sS}(\vx)
\left(
\mK_{\sS\sS}
+
\lambda_{\mathrm{ntk}}\mI
\right)^{-1}
\vy_{\sS}
,
\label{eq:ntk-predictor}
\end{equation}
where $\lambda_{\mathrm{ntk}}>0$ is a ridge regularization coefficient.

In the infinite-width and linearized-training regime, the NTK predictor characterizes the behavior of a gradient-trained neural network~\citep{NEURIPS2019_0d1a9651}. In practical
finite-width networks, however, nonlinear feature learning, stochastic optimization, and finite-kernel approximations generally cause the NTK prediction to differ from the prediction of the corresponding trained neural model~\citep{NEURIPS2020_ad086f59}. For a model trained on $\sS$, this discrepancy can be expressed as
\begin{equation}
\mR_t^{\sS}
=
f_{\vtheta_{\sS}^{*}}
\left(
\vx_t
\right)
-
g_{\sS}
\left(
\vx_t
\right).
\label{eq:model-ntk-gap}
\end{equation}

%% file: 4_method.tex
\begin{figure}[t]
    \centering
    \includegraphics[width=\linewidth]{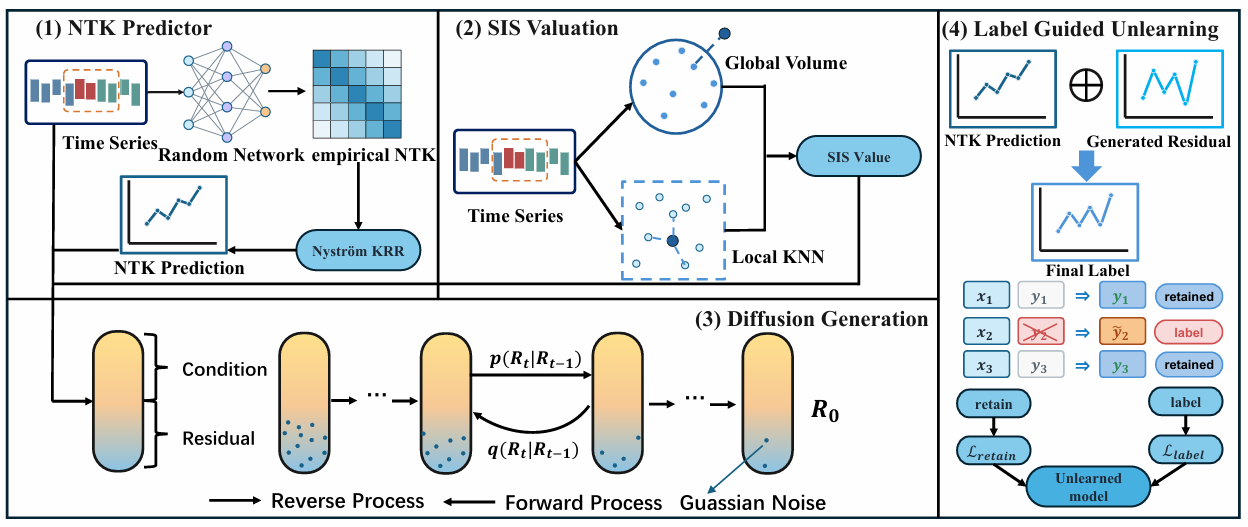}
    \vspace{-8mm}
    \caption{Overview of our proposed time-series unlearning framework RDTU. (1) The retained-set NTK predictor produces a base prediction for the affected window. (2) SIS integrates global directional contribution and local retained support to quantify the structural irreplaceability. (3) Conditioned on the NTK prediction and SIS, the residual diffusion model generates a residual correction. (4) The final label guides the unlearning update.}
    \label{fig:model-framework}
\end{figure}

\textbf{Proposed Approach Overview.}
To achieve the approximate unlearning of time series, we propose \textsc{RDTU}, a
\underline{\textbf{R}}esidual
\underline{\textbf{D}}iffusion framework of \underline{\textbf{T}}ime-series \underline{\textbf{U}}nlearning. As illustrated in Figure~\ref{fig:model-framework}, the overall method consists of four main components.  First, a retained-set neural tangent kernel (NTK) predictor provides a deletion-compatible base prediction that enables an approximation of the target exact unlearning model. Second, the Structural Irreplaceability Score (SIS) complements the influence estimation by quantifying the global and local structural support provided by each closure window, thereby characterizing the extent to which the unlearned data are replaceable for prediction. Third, conditioned on the NTK prediction and the SIS-based valuation of the forgetting windows, a diffusion model learns the residual between the NTK prediction and the model prediction, generating a refined label field for unlearning. Finally, the generated label field guides the unlearning update, enabling effective removal of the forgetting windows while preserving forecasting performance on strictly retained windows.

\vspace{-5mm}
\subsection{NTK Predictor Based Label Construction}
\label{subsec:ntk-predictor}
Random labeling provides a practical route for approximate unlearning: rather than preserving the original targets of deleted samples, the unlearned
model is guided toward constructed targets that reflect the desired post-deletion behaviour. In categorical tasks, such targets are usually formed by replacing the ground-truth label with an alternative label or a prescribed distribution~\citep{chundawat2023can,di2026label}. This option is not directly available for time-series forecasting, where each target is a continuous future trajectory whose value is correlated to its temporal context. Assigning an arbitrary value, such as zero or a retained target, does not in general represent the prediction of a model that has never been trained on the deleted interval.

Follow the definition of approaximate unlearning in Section~\ref{subsec:strict-unlearning}, the ideal label for each closure window should be the prediction
of the exact-unlearned model. Let
$\vtheta_{\mathrm{r}}^{*}$ denote the parameters obtained by training the forecasting model from scratch on the strict retained set $\train_{\mathrm{r}}$. For
each closure query $(\vx_t,\vy_t)\in\train_{\mathrm{f}}$, the ideal target is $\vy_t^{*}
=
f_{\vtheta_{\mathrm{r}}^{*}}
\left(
\vx_t
\right).$

Obtaining $\vy_t^{*}$ from exact unlearned model incurs the substantial computational cost. Therefore, we constructed an NTK predictor as a basic predictor that can be processed, in order to closely approximate our target model with lower cost.

Let \(\vtheta^{(n)}\), \(n=1,\ldots,N\), be \(N\) independently sampled initial parameter vectors of the forecasting architecture. For each initialization,
we define the empirical neural tangent kernel as
\begin{equation}
K^{(n)}
\left(
\vx,
\vx'
\right)
=
\nabla_{\vtheta}
f_{\vtheta^{(n)}}
\left(
\vx
\right)^{\top}
\nabla_{\vtheta}
f_{\vtheta^{(n)}}
\left(
\vx'
\right).
\label{eq:empirical-ntk}
\end{equation}
This kernel measures the similarity between two input windows through the alignment of their parameter gradients at the random initialization. Windows
with similar local parameter responses receive large kernel values and can therefore share predictive information through kernel regression.Then we construct a kernel ridge predictor around the random initialization output:
\begin{equation}
\vg_{\sS}^{(n)}
\left(
\vx
\right)
=
f_{\vtheta^{(n)}}
\left(
\vx
\right)
+
\vk_{\sS}^{(n)}
\left(
\vx
\right)
\left(
\mK_{\sS\sS}^{(n)}
+
\lambda_{\mathrm{ntk}}\mI
\right)^{-1}
\left(
\vy_{\sS}
-
f_{\vtheta^{(n)} \sS}
\right).
\label{eq:ntk-krr-predictor}
\end{equation}
In Equation~\ref{eq:ntk-krr-predictor}, \(\mK_{\sS\sS}^{(n)}\) and \(\vk_{\sS}^{(n)}(\vx)\)
encode similarities among the available windows and between the query and those windows, respectively. Here, \(\mI\) is an identity matrix matching the kernel
matrix. The vectors \(\vy_{\sS}\) and \(\vf_{\vtheta^{(n)}\sS}\) stack vectorized
targets and predictions \(\vf_{\vtheta^{(n)}}(\vx_t)\) in the same sample order.
Their difference is the initialization residual. The ridge coefficient stabilizes the regression. For computational efficiency, we approximate these kernel quantities using Nyström features~\citep{williams2000using}, avoiding construction of the full empirical NTK matrix while
preserving the same kernel-ridge objective. We further average predictors from multiple independent random initializations to reduce initialization-specific
variation~\citep{lakshminarayanan2017simple}:
\begin{equation}
g_{\sS}
\left(
\vx
\right)
=
\frac{1}{N}
\sum_{n=1}^{N}
\vg_{\sS}^{(n)}
\left(
\vx
\right).
\label{eq:ensemble-ntk-predictor}
\end{equation}
Then we can constract NTK predictors of the original training set
\(\train\) or the strictly retained set \(\train_{\mathrm{r}}\) respectively. However, the NTK predictor provides only a linearized approximation of neural-network training. It cannot fully capture finite-width feature learning, nonlinear parameter interactions, or the stochastic optimization dynamics of the actual model~\citep{chizat2019lazy,yang2021tensor}. Consequently, its predictions exhibit a substantial and non-negligible discrepancy from those of the trained neural predictor. For the original model and its corresponding NTK predictor, we define this prediction gap as the residual:
\begin{equation}
\mR_t^{\train}
=
f_{\vtheta^*}
\left(
\vx_t
\right)
-
g_{\train}
\left(
\vx_t
\right).
\label{eq:source-residual-target}
\end{equation}
A straightforward approach to residual correction is to fit a deterministic regressor between the available conditions and the residuals. However, such methods impose a restricted functional form and generally produce only a point estimate, which may be insufficient for capturing nonlinear or heteroscedastic. We therefore employ a conditional diffusion model to learn the distribution of residual corrections. Importantly, this construction is agnostic to the architecture of the original forecasting model. It does not require access to model-specific hidden representations, gradients, or parameter structures, but relies on the NTK prediction and the features of the corresponding data. 

The source residual defined in Equation~\ref{eq:source-residual-target} provides supervision for this
process, enabling the diffusion model to capture the finite-width discrepancy between the trained neural predictor and its NTK approximation. Because a deletion request typically removes only a small portion of the training data, we assume that it induces a limited shift in the overall temporal-pattern distribution. Consequently, the residual patterns learned from the complete training set remain transferable to the strictly retained-set conditions, and thus we can obtain the residuals $\widehat R_t^{D_r}$ on the retained dataset. For each closure window, the diffusion model receives the window
$\vx_t\in\mathbb{R^\ell}$ and the corresponding NTK prediction
$g(\vx_t)$ as conditional information: $\vc_t=[\vx_t,g(\vx_t)]$.

% \begin{equation}
% \vc_t
% =
% \left[
% x_\tau
% ,
% g
% \left(
% \vx_t
% \right)
% \right].
% \label{eq:diffusion-condition}
% \end{equation}
The forward variance schedule and cumulative product are defined as $\alpha_s = 1-\beta_s$ and
$\bar{\alpha}_s = \prod_{q=1}^{s}\alpha_q$ respectively.
Here, \(\bar{\vr}_t^{\train}
=(\operatorname{vec}(\mR_t^{\train})-\mu_r\vone)/\sigma_r\in\R^{hm}\)
is the standardized vectorized source residual. For a randomly sampled diffusion step, Gaussian noise is added to the clean source residual according to the predefined variance schedule:
\begin{equation}
\rvr_{t,s}
=
\sqrt{
\bar{\alpha}_s
}
\bar{\vr}_t^{\train}
+
\sqrt{
1-
\bar{\alpha}_s
}
\rvepsilon,
\qquad
\rvepsilon
\sim
\mathcal{N}
\left(
\vzero,
\mI
\right).
\label{eq:diffusion-forward-process}
\end{equation}
The denoiser receives the noisy residual, the diffusion step, and the conditional information, and is trained to predict the injected noise:
\begin{equation}
\Ls_{\mathrm{diff}}
=
\E
\left[
\left\|
\rvepsilon
-
\bm{\epsilon}_{\bm{\psi}}
\left(
\rvr_{t,s},
s,
\vc_t
\right)
\right\|_2^2
\right].
\label{eq:diffusion-noise-loss}
\end{equation}
\subsection{Structural Irreplaceability Score Valuation}
\label{subsec}

Training diffusion on raw data is highly expensive, with costs comparable to retraining on the retained set. Using raw data as the conditional input for the diffusion model is to extract the features of the corresponding data and generate the corresponding residual values. Since the NTK prediction values have to some extent already reflected the impact of this data on the entire model, we introduce the Structural
Irreplaceability Score (SIS), a compact reference-based descriptor that quantifies how difficult each closure query is to replace using the available reference patterns. SIS combines a global criterion and a local criterion to measure the replaceability of data.

We quantify global replaceability by measuring the change of Volume value in the directional structure of the reference dataset induced by a closure query.
Each forecasting window is first converted into a standardized vector
representation  $\bm{\phi}_t=\vx_t$.

We stack the representations of all windows in the current reference dataset $\sS$ into the reference matrix $\mPhi_{\sS}$. Its regularized second-moment matrix summarizes the directions covered by the corresponding temporal-pattern space:
\begin{equation}
\mG_{\sS}
=
\mPhi_{\sS}^{\top}
\mPhi_{\sS}
+
\lambda_{v,\sS}
\mI_{d_x},
\qquad
\lambda_{v,\sS}
=
\epsilon_v
\frac{
\operatorname{Tr}
\left(
\mPhi_{\sS}^{\top}
\mPhi_{\sS}
\right)
}{
d_x
}.
\label{eq:SIS-reference-gram}
\end{equation}
The log-determinant of the phase-dependent reference matrix measures the volume covered by the corresponding temporal-pattern space. For a closure query, we define its global contribution as the absolute volume change caused by removing its representation from the complete training set or adding its representation to the strictly retained set:
\begin{equation}
v_t^{\mathrm{global}}
=
\frac{1}{2}
\left|
\log\det
\left(
\mG_{\sS}
+
\eta_{\sS}
\bm{\phi}_t
\bm{\phi}_t^{\top}
\right)
-
\log\det
\left(
\mG_{\sS}
\right)
\right|
=
\frac{1}{2}
\left|
\log
\left(
1+
\eta_{\sS}
\bm{\phi}_t^{\top}
\mG_{\sS}^{-1}
\bm{\phi}_t
\right)
\right|.
\label{eq:SIS-global-volume}
\end{equation} 

The global volume quantifies the coverage of the temporal-pattern space as a whole. However, temporal dependence is often locally structured: retained
windows with similar recent contexts can provide substantially stronger support for a query than patterns represented only at the global level. Global coverage alone may therefore fail to reveal whether a query has close counterparts. We consequently measure local support by the average KNN distance to the nearest windows:
\begin{equation}
v_t^{\mathrm{local}}
=
\left(
\frac{1}{k}
\sum_{j\in\sJ_k(t;\sS)}
\left\|
\bm{\phi}_t
-
\bm{\phi}_{j}
\right\|_2^2
\right)^{1/2}.
\label{eq:SIS-local-isolation}
\end{equation}
We standardize the global and local scores and  then combine them into a single SIS value:
\begin{equation}
s_t
=
\beta
\mathrm{Norm}_{\sS}
\left(
v_t^{\mathrm{global}}
\right)
+
\left(
1-\beta
\right)
\mathrm{Norm}_{\sS}
\left(
v_t^{\mathrm{local}}
\right),
\qquad
0 \leq \beta \leq 1.
\label{eq:SIS-condition}
\end{equation}
The coefficient controls the relative contribution of global non-redundancy and local isolation, and we use equal weights by default. A large SIS value indicates that the query is poorly supported by the strict retained set. Then, for each closure window, the conditional information of the diffusion model will be changed to:
\begin{equation}
\vc_t
=
\left[
\vx_t
,
g
\left(
\vx_t
\right),
s_t
\right].
\label{eq:diffusion-condition}
\end{equation}
At unlearning time, reverse diffusion generates residual samples
$\left\{\hat{\bar{\vr}}_t^{(m)}\right\}_{m=1}^{M}$
from the conditional distribution
$p_{\bm{\psi}}\left(\bar{\vr}\mid\vc_t\right)$, using the available information corresponding to the strictly reserved set. We average the generated residual samples and restore the original residual scale:
\begin{equation}
\widehat R_t^{D_r}
=
\operatorname{mat}_{h,p}\!\left(
\mu_r\vone
+
\sigma_r
\frac{1}{M}
\sum_{m=1}^{M}
\hat{\bar{\vr}}_t^{(m)}
\right).
\label{eq:generated-residual}
\end{equation}
The final label for each affected window is then obtained by adding the generated residual correction to the retained-set NTK prediction:
\begin{equation}
\tilde{\vy}_t
=
g_{\train_{\mathrm{r}}}
\left(
\vx_t
\right)
+
\widehat R_t^{D_r}.
\label{eq:counterfactual-pseudolabel}
\end{equation}
\subsection{Label Field-Guided Unlearning}
\label{subsec:pseudolabel-unlearning}

The diffusion model produces the label
$\tilde{\vy}_t$ for every closure window. Updating the model with such constructed targets is conceptually aligned with label-based and teacher-based unlearning strategies, which redirect the predictions of affected samples while preserving useful knowledge learned from the retained data
~\citep{chundawat2023can,NEURIPS2023_062d711f,graves2021amnesiac}. In our setting, the labels are generated from retained-set information and are used to update the original model.

We initialize the unlearned model from the original parameters $\vtheta_{\mathrm{u}}^{(0)}=\vtheta^*$. At each update step, we draw a shuffled mini-batch $\sM_{\mathrm{f}}\subset\train_{\mathrm{f}}$ of closure
queries and an independent random mini-batch
$\sM_{\mathrm{r}}\subset\train_{\mathrm{r}}$ of strictly retained supervised windows. The retained mini-batch provides an anchoring signal that stabilizes
the update and preserves forecasting behaviour on unaffected temporal patterns and the closure mini-batch is aligned with the generated labels and supplies the forgetting signal that redirects predictions away from their original training-induced behaviour. We define the retained anchoring loss and the label alignment loss as
\begin{equation}
\Ls_{\mathrm{r}}
=
\frac{1}
{\left|\sM_{\mathrm{r}}\right|}
\sum_{(\vx_t,\vy_t)\in\sM_{\mathrm{r}}}
\left\|
f_{\vtheta_{\mathrm{u}}}
\left(
\vx_t
\right)
-
\vy_t
\right\|_{\mathrm{F}}^2,\,\, \Ls_{\mathrm{f}}
=
\frac{1}
{\left|\sM_{\mathrm{f}}\right|}
\sum_{(\vx_t,\vy_t)\in\sM_{\mathrm{f}}}
\left\|
f_{\vtheta_{\mathrm{u}}}
\left(
\vx_t
\right)
-
\tilde{\vy}_t
\right\|_{\mathrm{F}}^2.
\label{eq:retained-anchoring-loss}
\end{equation}
The final update is computed according to the following objective:
\begin{equation}
\Ls_{\mathrm{U}}
=
\lambda_{\mathrm{r}}
\Ls_{\mathrm{r}}
+
\lambda_{\mathrm{f}}
\Ls_{\mathrm{f}}.
\label{eq:pseudolabel-unlearning-objective}
\end{equation}

%% file: 5_exp.tex
\subsection{Datasets and Experimental Setups}
\label{subsec:experimental_setups}
\textbf{Datasets.}
We conduct experiments on four public time-series datasets. Air Quality~\citep{beijing_multi-site_air_quality_501} and Traffic\citep{lai2018modeling} are used for univariate forecasting, whereas EEG\citep{zyma2019electroencephalograms} and HAR70+\citep{har70plus_780} are used for multivariate forecasting. For each dataset, we select 10 series or subjects and train a separate forecasting model for each one. Each model is evaluated under three deletion seeds, yielding 30 deletion requests per dataset. We first average the results across deletion seeds for each series or subject and then report the macro-average across the 10 series or subjects.

\textbf{Evaluation Metrics.} We evaluate unlearning effectiveness using membership indistinguishability ~\citep{shokri2017membership,carlini2022membership} and functional agreement with exact retraining. The former is measured by the two-sided U/N AUC, where values closer to $0.5$ indicate stronger forgetting. The latter is measured by the F-P Gap, namely the normalized RMSE between unlearned and exact-retrained predictions on forgetting inputs, reported as a percentage. Forecasting utility is evaluated using test NRMSE, while computational efficiency is measured by wall-clock time. Detailed metric and timing implementations are provided in the Appendix.

\textbf{Baselines.}
We compare our model against a sizable collection of representative baselines: \textbf{SSD}~\citep{foster2024fast}, \textbf{NegGrad+}~\citep{NEURIPS2023_062d711f}, \textbf{Random label}, \textbf{Zero label}, \textbf{SCRUB-R}~\citep{NEURIPS2023_062d711f}, \textbf{TS Unlearn}~\citep{wang2026ts}, and we use the \textbf{Transformer}~\citep{vaswani2017attention} as our base model. The details are described in the Appendix.
\subsection{Overall Performance}

\label{subsec:overall Performance}
\begin{table*}[h]
\centering
\caption{
Performance comparison between our method and other unlearning baselines on univariate and multivariate forecasting tasks. Unlearning effectiveness is evaluated by the two-sided U/N AUC and the F-P Gap(\%), where U/N values closer to $0.5$ indicate lower distinguishability between forgotten and never-seen samples, and a lower F-P Gap(\%) indicates closer predictive agreement with exact retraining. Forecasting utility is evaluated by test NRMSE, with lower values indicating better predictive performance. Computational efficiency is measured by wall-clock time in seconds.
}
\label{tab:main_results}

\setlength{\tabcolsep}{2.5pt}
\renewcommand{\arraystretch}{1.15}

\resizebox{\textwidth}{!}{%
\begin{tabular}{l | cccc | cccc | cccc | cccc}
\hline

\multirow{3}{*}{\textbf{Method}}
& \multicolumn{8}{c|}{\textbf{Univariate Forecasting}}
& \multicolumn{8}{c}{\textbf{Multivariate Forecasting}} \\

\cline{2-17}

& \multicolumn{4}{c|}{\textbf{Air Quality}}
& \multicolumn{4}{c|}{\textbf{Traffic}}
& \multicolumn{4}{c|}{\textbf{EEG}}
& \multicolumn{4}{c}{\textbf{HAR70}} \\

\cline{2-17}

& $\mathbf{U/N}$
& \shortstack{\textbf{F-P Gap}}
& \textbf{NRMSE}
& \textbf{Time}

& $\mathbf{U/N}$
& \shortstack{\textbf{F-P Gap}}
& \textbf{NRMSE}
& \textbf{Time}

& $\mathbf{U/N}$
& \shortstack{\textbf{F-P Gap}}
& \textbf{NRMSE}
& \textbf{Time}

& $\mathbf{U/N}$
& \shortstack{\textbf{F-P Gap}}
& \textbf{NRMSE}
& \textbf{Time} \\

\hline

Original
& 0.6009 & 6.56 & 0.2546 & --
& 0.6454 &15.57 & 0.3941 & --
& 0.6389 & 5.24 & 0.1069 & --
& 0.6911 & 11.96 & 0.3495 & -- \\

Retrain
& 0.5034 & -- & 0.2622 & 26.323
& 0.5291 & -- & 0.3955 & 12.549
& 0.5235 & -- & 0.1148 & 8.910
& 0.5210 & -- & 0.3459 & 32.371 \\

\hline

SSD
& 0.5218 & 52.8954 & 0.6683 & 0.946
& 0.5371 & 72.6908 & 0.8928 & 0.395
& 0.5180 & 54.1037 & 0.5877 & 0.227
& 0.5204 & 72.0281  & 0.5763 & 0.623 \\

NegGrad+
& 0.5090 & 162.1498 & 1.5902 & 9.871
& 0.5412 & 140.9170 & 1.3989 & 4.340
& 0.5261 & 102.1933 & 0.9080 & 2.397
& 0.5372 & 288.0736 & 1.6082 & 10.051 \\

Zero-label
& 0.5093 & 54.0268 & 0.6009 & 4.161
& 0.5387 & 51.8856 & 0.6896 & 2.150
& 0.5184 & 60.1355 & 0.5854 & 0.698
& 0.5227 & 58.6257  & 0.4214 & 1.009 \\

Random-label
& 0.5069 & 54.2146 & 0.6052 & 0.775
& 0.5374 & 52.5161 & 0.7063 & 0.786
& 0.5183 & 60.0174 & 0.5846 & 0.695
& 0.5220 & 45.3235 & 0.5377  & 1.007 \\

SCRUB-R
& 0.5195 & 8.4768 & 0.2720 & 6.854
& 0.5499 & 11.9866 & 0.4142 & 4.913
& 0.5368 & 18.9612 & 0.2149 & 2.992
& 0.5497 & 11.6701 & 0.3565  & 4.802 \\

TS-Unlearn
& 0.5197 & 7.5967 & 0.2698 & 8.227
& 0.5458 & 13.5214 & 0.4189 & 3.579
& 0.5254 & 8.5701 & 0.1353 & 1.310
& 0.5385 & 12.8356 & 0.3539 & 1.933 \\

\hline

\textbf{Ours}
& \textbf{0.5001} & \textbf{4.6312} & \textbf{0.2600} & \textbf{11.710}
& \textbf{0.5420} & \textbf{9.9251} & \textbf{0.3935} & \textbf{4.261}
& \textbf{0.5244} & \textbf{3.7526} & \textbf{0.1110} & \textbf{2.322}
& \textbf{0.5176} & \textbf{9.7703} & \textbf{0.3424} & \textbf{5.744} \\

\hline
\end{tabular}%
}
\end{table*}
\vspace{-3mm}
Table~\ref{tab:main_results} present the overall performance comparison of RDTU against all baselines across the four time-series datasets. \textbf{\textit{Our RDTU consistently and significantly outperforms all competing methods under evaluation metrics of F-P Gap and NRMSE across all datasets.}} Specifically, RDTU achieves substantial improvements over the best baselines, SCRUB-R and TS-Unlearn, with relative gains ranging from 16.3\% to 56.2\% in F-P Gap and 3.2\% to 18.0\% in NRMSE. Although our performance on the U/N auc metric was not completely optimal, it can be observed that methods such as NegGrad+ and Random-label only achieved a slight advantage on U/N, but led to a complete deterioration in other metrics. This result is consistent with the limitations discussed earlier: aggressive gradient reversal or arbitrary target replacement may reduce membership distinguishability by broadly disrupting the learned forecasting function, rather than selectively removing the influence of the forgotten data. \textbf{\textit{Compared to other forgetting models that did not experience training breakdown, we still maintained a good advantage on U/N auc, which indicates that we achieved excellent forgetting effects on these datasets.}} Extended experiments of different forecasting horizons shows in the Appendix.

\subsection{Ablation Study}
\label{subsec:ablation Study}
We conduct a systematic ablation study to investigate the contribution of each key component of our RDTU with the following ablated variants: 
\textbf{w/o diffusion} means we conduct the experiments without residual calculation; the \textbf{residual-MLP} and \textbf{residual-ARIMA} means we conduct the experiments with other methods such as MLP and ARIMA to get the residual; \textbf{w/o SIS} removes the SIS from the diffusion condition and generates residuals using only the NTK prediction.
\vspace{-5mm}
\begin{table*}[h]
\centering
\caption{
Ablation study of the key components of our method on different datasets.
}
\label{tab:ablation_results}
\setlength{\tabcolsep}{2pt}
\renewcommand{\arraystretch}{1.15}

\resizebox{\textwidth}{!}{%
\begin{tabular}{l | ccc | ccc | ccc | ccc}
\hline

\multirow{3}{*}{\textbf{Methods}}
& \multicolumn{6}{c|}{\textbf{Univariate Forecasting}}
& \multicolumn{6}{c}{\textbf{Multivariate Forecasting}} \\

\cline{2-13}

& \multicolumn{3}{c|}{\textbf{Air Quality}}
& \multicolumn{3}{c|}{\textbf{Traffic}}
& \multicolumn{3}{c|}{\textbf{EEG}}
& \multicolumn{3}{c}{\textbf{HAR70}} \\

\cline{2-13}

& $\mathbf{U/N}$
& \shortstack{\textbf{F-P Gap}}
& \textbf{NRMSE}

& $\mathbf{U/N}$
& \shortstack{\textbf{F-P Gap}}
& \textbf{NRMSE}

& $\mathbf{U/N}$
& \shortstack{\textbf{F-P Gap}}
& \textbf{NRMSE}

& $\mathbf{U/N}$
& \shortstack{\textbf{F-P Gap}}
& \textbf{NRMSE} \\

\hline

w/o Diffusion
& 0.5250 & 5.2378 & 0.2685
& 0.5455 & 11.3759 & 0.3946
& 0.5297 & 4.2684& 0.1232 
& 0.5190 & 9.8371 & 0.3434 \\

Residual-MLP
& 0.5201 & 5.0807 & 0.2632
& 0.5454 & 10.3245 & 0.3940
& 0.5341 & 5.5752 & 0.2583
& 0.5261 & 10.064 & 0.3398\\

Residual-ARIMA
  & 0.5392 & 4.9284 & 0.2609
& 0.5381& 10.7915 & 0.3968
& 0.5218 & 4.2113 &0.1135
& 0.5175 & 10.8921 & 0.3498 \\

w/o SIS
& 0.5180 & 5.0545& 0.2610
& 0.5475 & 10.2339 & 0.3964
& 0.5226 & 3.9427 & 0.1171
& 0.5213 & 10.0836 & 0.3452\\

\hline

\textbf{Ours}
& \textbf{0.5001} & \textbf{4.6312} & \textbf{0.2600} 
& \textbf{0.5420} & \textbf{9.9251} & \textbf{0.3935} 
& \textbf{0.5244} & \textbf{3.7526} & \textbf{0.1110} 
& \textbf{0.5176} & \textbf{9.7703} & \textbf{0.3424}  \\

\hline
\end{tabular}%
}
\end{table*}
The results and comparison of these variants are shown in Table~\ref{tab:ablation_results}. \textbf{\textit{Residual-related ablations show that removing residual correction degrades the quality of the constructed labels, confirming that the gap between the NTK predictor and the forecasting model cannot be ignored.}} Replacing diffusion with a residual MLP or ARIMA also yields inferior results, as deterministic regression may fail to capture the complex residual distribution. These results validate both residual correction and diffusion-based residual modeling. Removing SIS increases the F-P Gap by 3.1\%-9.1\%, confirming that structural information complements NTK predictions in residual generation.

\subsection{Hyperparameter Analysis}
\label{subsec:hyperparameter}
\begin{figure}[h]
    \centering
    \includegraphics[width=\linewidth]{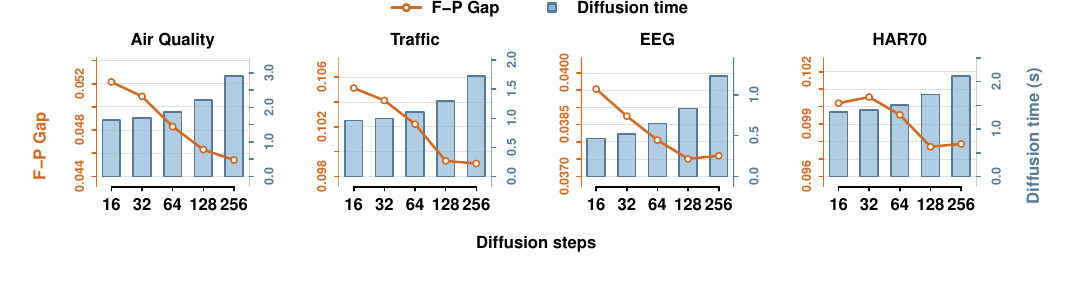}
    \vspace{-10mm}
    \caption{
    Sensitivity analysis of RDTU under different diffusion steps of Horizon Length=1.
    }
    \label{fig:step_1}
\end{figure}
We vary the number of diffusion steps while keeping all other settings fixed, and report the F-P Gap and diffusion-only runtime on different datasets. As shown in Figure~\ref{fig:step_1}, increasing the number of steps from 16 to 128 reduces the F-P Gap by 0.8\% to 11.2\%. Increasing the number of steps further to 256 results in only marginal changes of -0.01to 0.09 percentage points, while increasing the diffusion time by 0.39 to 0.70 seconds. \textbf{\textit{These results indicate that insufficient denoising steps limit residual-generation quality, whereas excessive steps introduce additional cost with diminishing benefits.}} We therefore use $steps=128$ as the default setting.

%% file: 6_Appendix.tex
\section{More Implementation Details}
\subsection{Datasets}

We conduct experiments on four public time-series datasets. Air Quality and Traffic are used for univariate forecasting, whereas EEG and HAR70+ are used for multivariate forecasting. The forecasting horizon is one time step for all datasets. For each dataset, we select 10 series or subjects and train a separate forecasting model for each one. Each model is evaluated under three deletion seeds, yielding 30 deletion requests per dataset. We first average the results across deletion seeds for each series or subject and then report the macro-average across the 10 series or subjects.A deletion request is specified as a contiguous block of raw observations. We remove every forecasting example whose input or prediction target overlaps this block, thereby enforcing strict temporal deletion closure. Below, the input-window length denotes the number of historical observations provided to the forecasting model, while the deletion-block length denotes the number of consecutive raw observations selected for forgetting.

\begin{itemize}[leftmargin=*]

\item \textit{Air Quality}:
This dataset contains hourly air-pollutant and meteorological measurements collected from 12 monitoring sites in Beijing between March 2013 and February 2017 \citep{beijing_multi-site_air_quality_501}. We treat the PM$_{2.5}$ measurements from each site as an independent univariate time series and predict its concentration at the next time step. We use an input window of 24 hourly observations.

\item \textit{Traffic}:
This dataset contains hourly road occupancy measurements from 862 freeway sensors in the San Francisco Bay Area during 2015--2016 \citep{lai2018modeling}. We treat each sensor as an independent univariate time series and predict its occupancy rate at the next time step. We use an input window of 16 hourly observations.

\item \textit{EEG}:
The EEGMAT dataset contains 19-channel EEG recordings from 36 subjects performing mental arithmetic tasks, sampled at 500 Hz \citep{zyma2019electroencephalograms}. We use the Fp1, Fp2, and F3 channels and jointly predict their values at the next time step. We use an input window of 32 samples.

\item \textit{HAR70+}:
This dataset contains accelerometer recordings from 18 older adults wearing tri-axial sensors on the lower back and thigh, sampled at 50 Hz \citep{har70plus_780}. We use the \texttt{back\_y}, \texttt{thigh\_x}, and \texttt{thigh\_y} channels and jointly predict their values at the next time step. We use an input window of 32 samples.

\end{itemize}

\subsection{Phase-Specific SIS Construction}
\label{app:phase-specific-sis}

SIS is used in both diffusion training and counterfactual residual generation. Although the two phases share the same scoring framework, they evaluate opposite structural perturbations with respect to different reference sets. During diffusion training, the complete training set $\train$ serves as the reference, and SIS measures the structural change caused by removing the query. During residual generation, the strictly retained set $\train_{\mathrm r}$ becomes the reference, and SIS measures the change caused by adding the query. We represent these two operations using a phase-dependent reference set $\sS\in{\train,\train_{\mathrm r}}$ and direction indicator. 
\begin{equation}
\eta_{\sS}
=
\begin{cases}
-1, & \sS=\train,\\
+1, & \sS=\train_{\mathrm r}.
\end{cases}
\label{eq:app-sis-direction}
\end{equation}

During diffusion training, $\sS=\train$ and the global SIS component measures
the structural change caused by removing the query from the complete training
set. During residual generation, $\sS=\train_{\mathrm r}$ and it measures the
change caused by adding the query to the strictly retained set.

For each query window, let $\bm{\phi}t=\vx_t$ denote its vector representation, and let $\mPhi{\sS}$ stack the representations of all windows in the reference set. The structural geometry of $\sS$ is summarized by the following regularized second-moment matrix:

\begin{equation}
\mG_{\sS}
=
\mPhi_{\sS}^{\top}\mPhi_{\sS}
+
\lambda_{v,\sS}\mI_{d_x},
\qquad
\lambda_{v,\sS}
=
\epsilon_v
\frac{
\operatorname{Tr}
\left(
\mPhi_{\sS}^{\top}\mPhi_{\sS}
\right)
}{
d_x
}.
\label{eq:app-sis-reference-matrix}
\end{equation}

Here, $d_x$ is the representation dimension, while the adaptive ridge coefficient $\lambda_{v,\sS}$ stabilizes the matrix operations across reference sets with different scales. Based on this matrix, the global SIS component measures the absolute change in the log-determinant volume induced by the phase-specific perturbation:

\begin{equation}
v_{t,\sS}^{\mathrm{global}}
=
\frac{1}{2}
\left|
\log\det
\left(
\mG_{\sS}
+
\eta_{\sS}
\bm{\phi}_t\bm{\phi}_t^{\top}
\right)
-
\log\det
\left(
\mG_{\sS}
\right)
\right|.
\label{eq:app-sis-global}
\end{equation}
The global component captures overall directional change but does not indicate whether similar windows remain locally available. We therefore complement it with a nearest-neighbor measure of local support. To prevent the query from becoming its own nearest neighbor during diffusion training, the local reference set is defined as
\begin{equation}
\sS_{t}^{\mathrm{local}}
=
\begin{cases}
\train\setminus\{(\vx_t,\vy_t)\},
& \sS=\train,\\
\train_{\mathrm r},
& \sS=\train_{\mathrm r}.
\end{cases}
\label{eq:app-sis-local-reference}
\end{equation}

Let $\sJ_k(t;\sS_t^{\mathrm{local}})$ denote the indices of the $k$ nearest reference windows to query $t$. The local isolation score is then computed as

\begin{equation}
v_{t,\sS}^{\mathrm{local}}
=
\left(
\frac{1}{k}
\sum_{j\in\sJ_k(t;\sS_t^{\mathrm{local}})}
\left\|
\bm{\phi}_t-\bm{\phi}_j
\right\|_2^2
\right)^{1/2}.
\label{eq:app-sis-local}
\end{equation}
The resulting phase-specific SIS is
\begin{equation}
s_t^{\sS}
=
\beta
\operatorname{Norm}_{\sS}
\left(
v_{t,\sS}^{\mathrm{global}}
\right)
+
(1-\beta)
\operatorname{Norm}_{\sS}
\left(
v_{t,\sS}^{\mathrm{local}}
\right).
\label{eq:app-phase-specific-sis}
\end{equation}
Therefore, $s_t^{\train}$ characterizes the structural role of the query in the
complete training set and is used to train the diffusion model, whereas
$s_t^{\train_{\mathrm r}}$ characterizes its support under the retained-data
condition and is used for counterfactual residual generation.

\subsection{Conditional Residual Diffusion}
\label{app:conditional-residual-diffusion}

RDTU does not directly apply diffusion to the forecasting targets. Instead, it
models the discrepancy between a trained forecasting model and its corresponding
NTK predictor. This design separates the counterfactual label into an
NTK-based prediction and a residual correction. The NTK predictor provides the
principal prediction under a given reference dataset, while the diffusion model
captures the finite-width and nonlinear effects that cannot be represented by
the NTK approximation.

\paragraph{Residual diffusion training.}
The diffusion model is trained under the complete-data condition. For each
training window $\vx_t$, we compute the source residual between the original
forecasting model and the complete-data NTK predictor. Because residuals from
different datasets and forecasting dimensions may have different scales, they
are standardized before diffusion training:
\begin{equation}
\begin{aligned}
\mR_t^{\train}=
f_{\vtheta_{\train}^{*}}
\left(
\vx_t
\right)
-
g_{\train}
\left(
\vx_t
\right),
&&\bar{\vr}_t^{\train}=
\frac{
\operatorname{vec}
\left(
\mR_t^{\train}
\right)
-
\mu_r\vone
}{
\sigma_r
}.
\end{aligned}
\label{eq:app-source-residual}
\end{equation}
Residual generation depends on both the model-based prediction and the
structural role of the query in the reference dataset. We therefore condition
the denoiser on the input window, the complete-data NTK prediction, and the
complete-data SIS:
\begin{equation}
\vc_t^{\train}
=
\left[
\vx_t^{\top},
g_{\train}
\left(
\vx_t
\right)^{\top},
s_t^{\train}
\right]^{\top}.
\label{eq:app-diffusion-training-condition}
\end{equation}
Here, $g_{\train}(\vx_t)$ represents the prediction induced by the complete
training set, while $s_t^{\train}$ measures the structural change caused by
removing the query from that set. The query window $\vx_t$ retains its temporal
context, allowing the diffusion model to distinguish residual patterns
associated with different temporal inputs.

Let $\{\beta_s\}_{s=1}^{S_{\mathrm{diff}}}$ denote the diffusion variance
schedule, with $\alpha_s=1-\beta_s$ and
$\bar{\alpha}_s=\prod_{q=1}^{s}\alpha_q$. During training, a diffusion step $s$
and Gaussian noise $\rvepsilon$ are sampled, the standardized source
residual is perturbed according to the forward diffusion process. The denoiser
is trained to recover the injected noise:
\begin{equation}
\begin{aligned}
\rvr_{t,s}
=
\sqrt{\bar{\alpha}_s}
\bar{\vr}_t^{\train}
+
\sqrt{1-\bar{\alpha}_s}
\rvepsilon,
&&
\rvepsilon
\sim
\mathcal{N}
\left(
\vzero,\mI
\right),
&&\Ls_{\mathrm{diff}}=
\E_{t,s,\rvepsilon}
\left[
\left\|
\rvepsilon
-
\bm{\epsilon}_{\bm{\psi}}
\left(
\rvr_{t,s},
s,
\vc_t^{\train}
\right)
\right\|_2^2
\right].
\end{aligned}
\label{eq:app-diffusion-training}
\end{equation}
Through this objective, the denoiser learns the conditional distribution of the
residual discrepancy rather than a deterministic point estimate. This is
particularly useful when similar NTK predictions correspond to different
residual corrections because of nonlinear feature learning, stochastic
optimization, or local temporal variation.

\paragraph{Transfer to the retained-data condition.}
At unlearning time, the trained denoiser is kept fixed. Because a deletion
request typically affects only a limited portion of the complete training set,
we assume that the residual relationship learned from the complete-data model
remains transferable after replacing the complete-data condition with the
retained-data condition. Importantly, the transferred object is the conditional
residual pattern, rather than the original complete-data prediction.

At unlearning time, for each affected closure query, we replace the complete-data NTK prediction and SIS with their retained-set counterparts to construct the counterfactual condition:
\begin{equation}
\vc_t^{\train_{\mathrm r}}
=
\left[
\vx_t^{\top},
g_{\train_{\mathrm r}}
\left(
\vx_t
\right)^{\top},
s_t^{\train_{\mathrm r}}
\right]^{\top}.
\label{eq:app-diffusion-generation-condition}
\end{equation}
The input window remains unchanged because it specifies the query for which the
counterfactual label is required. In contrast,
$g_{\train_{\mathrm r}}(\vx_t)$ represents the prediction supported by the
strictly retained set, and $s_t^{\train_{\mathrm r}}$ measures the structural
change caused by adding the closure query to that set. The difference between
$\vc_t^{\train}$ and $\vc_t^{\train_{\mathrm r}}$ therefore informs the
diffusion model about the change from the original-data condition to the
counterfactual retained-data condition.

\paragraph{Reverse residual generation.}
For each closure query, reverse diffusion begins with an independently sampled
Gaussian vector
$\hat{\bar{\vr}}_{t,S_{\mathrm{diff}}}^{(m)}$. At each step, the denoiser predicts the noise under the retained-data condition and progressively
reconstructs a standardized residual sample:
\begin{equation}
\begin{aligned}
\hat{\bar{\vr}}_{t,S_{\mathrm{diff}}}^{(m)}\sim
\mathcal{N}
\left(
\vzero,\mI
\right),
&&\hat{\bar{\vr}}_{t,s-1}^{(m)}=
\frac{1}{\sqrt{\alpha_s}}
\left[
\hat{\bar{\vr}}_{t,s}^{(m)}
-
\frac{\beta_s}{\sqrt{1-\bar{\alpha}_s}}
\bm{\epsilon}_{\bm{\psi}}
\left(
\hat{\bar{\vr}}_{t,s}^{(m)},
s,
\vc_t^{\train_{\mathrm r}}
\right)
\right]
+
\sqrt{\tilde{\beta}_s}\vz_s,
\end{aligned}
\label{eq:app-diffusion-reverse}
\end{equation}
Here, $\vz_s\sim\mathcal{N}(\vzero,\mI)$ for $s>1$, while
$\vz_1=\vzero$ in the final denoising step. We repeat the reverse process $M$
times and average the resulting samples to reduce sampling-specific variation.
The averaged residual is then restored to its original scale and added to the
retained-set NTK prediction:
\begin{equation}
\begin{aligned}
\widehat{\mR}_t^{\train_{\mathrm r}}
&=
\operatorname{vec}^{-1}
\left(
\mu_r\vone
+
\sigma_r
\frac{1}{M}
\sum_{m=1}^{M}
\hat{\bar{\vr}}_{t,0}^{(m)}
\right),
&&\tilde{\vy}_t=
g_{\train_{\mathrm r}}
\left(
\vx_t
\right)
+
\widehat{\mR}_t^{\train_{\mathrm r}}.
\end{aligned}
\label{eq:app-generated-residual-and-label}
\end{equation}
The resulting $\tilde{\vy}_t$ combines the retained-set prediction supplied by
the NTK predictor with a residual correction generated under the corresponding
retained structural condition. It is subsequently used as the guiding label
for updating the original forecasting model on the strict temporal deletion
closure.

\subsection{End-to-End Unlearning Procedure}
\label{app:end-to-end-procedure}

The complete procedure is summarized in
Algorithm~\ref{alg:rdtu}. Diffusion training can be performed once for the
original model, while each deletion request only requires constructing the
retained condition, generating the counterfactual labels, and updating the
forecasting model.

\begin{algorithm}[h]
\caption{\textsc{RDTU}: Residual-Diffusion-Guided Time-Series Unlearning}
\label{alg:rdtu}
\begin{algorithmic}[1]

\Require Original training set $\train$, original model
$f_{\vtheta_{\train}^{*}}$, and deletion request $\sT_{\mathrm f}$

\Ensure Unlearned model $f_{\vtheta_{\mathrm u}}$

\Statex \textbf{Complete-data diffusion training}

\State Construct the complete-data NTK predictor $g_{\train}$

\State Compute the source residuals $\mR_t^{\train}$ and complete-data
SIS values $s_t^{\train}$

\State Construct the conditions $\vc_t^{\train}$ and train
$\bm{\epsilon}_{\bm{\psi}}$ by minimizing $\Ls_{\mathrm{diff}}$

\Statex \textbf{Counterfactual label generation}

\State Construct the strict deletion closure
$\train_{\mathrm f}(\sT_{\mathrm f})$ and retained set
$\train_{\mathrm r}(\sT_{\mathrm f})$

\State Construct the retained-set NTK predictor $g_{\train_{\mathrm r}}$

\For{each $(\vx_t,\vy_t)\in\train_{\mathrm f}$}

    \State Compute the retained-set SIS $s_t^{\train_{\mathrm r}}$

    \State Generate
    $\widehat{\mR}_t^{\train_{\mathrm r}}$
    from
    $p_{\bm{\psi}}
    (\bar{\vr}\mid\vc_t^{\train_{\mathrm r}})$

    \State Construct
    $\tilde{\vy}_t
    \gets
    g_{\train_{\mathrm r}}(\vx_t)
    +
    \widehat{\mR}_t^{\train_{\mathrm r}}$

\EndFor

\Statex \textbf{Label field-guided unlearning}

\State Initialize
$\vtheta_{\mathrm u}^{(0)}
\gets
\vtheta_{\train}^{*}$

\State Optimize $\vtheta_{\mathrm u}$ on retained targets and generatedclosure labels using
$\lambda_{\mathrm{retain}}\Ls_{\mathrm r} + \lambda_{\mathrm{label}}\Ls_{\mathrm f}$

\State \textbf{return} $f_{\vtheta_{\mathrm u}}$

\end{algorithmic}
\end{algorithm}

\subsection{Metrics}
To evaluate \emph{unlearning effectiveness}, we consider both membership indistinguishability and functional agreement with exact retraining. Membership indistinguishability is measured by the two-sided U/N AUC, defined as $\max\{a,1-a\}$, where $a$ is the ROC AUC for distinguishing forgotten samples ($U$) from never-seen samples ($N$). A value closer to $0.5$ indicates that the forgotten samples are less distinguishable from non-members. Functional agreement is measured by the F-P Gap, defined as the RMSE between the predictions of the unlearned model and those of the exact-retraining model on the forgetting inputs, normalized by the standard deviation of the raw training targets and reported as a percentage. A lower F-P Gap indicates closer agreement with exact retraining. To evaluate \emph{forecasting utility}, we report the test NRMSE, computed as the test RMSE divided by the standard deviation of the raw training series before deletion, where a lower value indicates better predictive performance. \emph{Computational efficiency} is measured by the wall-clock time required to complete each unlearning method. For our method, the reported time includes NTK computation, diffusion training and sampling, and model updating.

\textbf{Membership Inference Attack} We conduct the membership audit at the level of temporal deletion units rather
than isolated observations. An audit candidate may therefore contain multiple
forecasting windows induced by the same temporal interval. Following the strict
temporal deletion closure defined in
Equation~\ref{eq:strict-closure}, we define the closure operator for an
arbitrary temporal interval $\sT'$ as
\begin{equation}
\mathcal{C}
\left(
\sT'
\right)
=
\left\{
(\vx_t,\vy_t)
\;\middle|\;
\sI_t\cap\sT'
\neq
\emptyset
\right\},
\qquad
\sI_t=[t-\ell,t+h).
\label{eq:audit-closure}
\end{equation}
Let $\sT_{\mathrm f}$ denote the raw timestamps included in a deletion request,
and let $\sT_{\mathrm n}$ denote an equally sized set of timestamps held out
before the original model is trained. The forgotten and never-seen audit
collections are defined as
\begin{equation}
\mathcal{A}_{\mathrm U}
=
\mathcal{C}
\left(
\sT_{\mathrm f}
\right),
\qquad
\mathcal{A}_{\mathrm N}
=
\mathcal{C}
\left(
\sT_{\mathrm n}
\right).
\label{eq:audit-un-sets}
\end{equation}
Thus, $\mathcal{A}_{\mathrm U}$ contains every forecasting window whose input
context or prediction target overlaps the requested deletion interval.
These windows were used to train the original model
$f_{\vtheta_{\train}^{*}}$ but should behave as non-members after unlearning.
In contrast, $\mathcal{A}_{\mathrm N}$ is excluded before the original model is
trained and therefore consists of genuine non-members.

The never-seen intervals are fixed for each user across deletion seeds, whereas
the forgotten intervals are independently sampled for each deletion request.
We additionally select a retained-member audit collection
$\mathcal{A}_{\mathrm M}\subset\train_{\mathrm r}$. To reduce distributional
bias, candidates in $\mathcal{A}_{\mathrm M}$ and
$\mathcal{A}_{\mathrm N}$ are matched to those in
$\mathcal{A}_{\mathrm U}$ using temporal descriptors and Hungarian matching.
The resulting M/U/N groups are mutually disjoint and contain the same number of
temporal audit units. For multivariate data, all variables observed within the
same temporal window are jointly treated as one forecasting sample rather than
as independent membership candidates.

\paragraph{Candidate-conditioned LiRA}
We implement membership inference using candidate-conditioned LiRA. For each
temporal audit unit $i$, shadow models are trained using the same background
training data while alternately including or excluding that candidate. Because
one candidate may induce multiple closure windows, the forecasting statistics
over all its associated windows are aggregated into a single attack-feature
vector $\vz_i(f)$. The feature vector contains squared error, absolute error,
signed error, residual variation, prediction statistics, and target statistics.

Using the outputs of the shadow models, we fit separate diagonal Gaussian
distributions to the observations obtained when candidate $i$ is included in
or excluded from shadow-model training:
\begin{equation}
p_{c,i}
=
\mathcal{N}
\left(
\bm{\mu}_{i}^{c},
\operatorname{diag}
\left(
\left(
\bm{\sigma}_{i}^{c}
\right)^2
\right)
\right),
\qquad
c\in
\left\{
\mathrm{in},
\mathrm{out}
\right\}.
\label{eq:lira-reference-distributions}
\end{equation}

For a target forecasting model $f_{\vtheta}$, the candidate-conditioned LiRA
score is the log-likelihood ratio

\begin{equation}
s_i^{\mathrm{LiRA}}
\left(
f_{\vtheta}
\right)
=
\log
p_{\mathrm{in},i}
\left(
\vz_i
\left(
f_{\vtheta}
\right)
\right)
-
\log
p_{\mathrm{out},i}
\left(
\vz_i
\left(
f_{\vtheta}
\right)
\right).
\label{eq:lira-score}
\end{equation}
A larger score indicates that the behavior of candidate $i$ is more consistent
with that of a training member. When auditing the original model
$f_{\vtheta_{\train}^{*}}$, both
$\mathcal{A}_{\mathrm M}$ and $\mathcal{A}_{\mathrm U}$ are treated as
members, whereas $\mathcal{A}_{\mathrm N}$ is treated as the non-member group.
When auditing the unlearned model $f_{\vtheta_{\mathrm u}}$, only
$\mathcal{A}_{\mathrm M}$ retain member-like behavior, while candidates
in both $\mathcal{A}_{\mathrm U}$ and $\mathcal{A}_{\mathrm N}$ should exhibit
non-member behavior. We use separate shadow-model distribution fits for the
original-model and unlearning audits.

\paragraph{Two-sided U/N AUC.}
Let $a_{\mathrm{U/N}}$ denote the conventional ROC AUC obtained by
distinguishing the LiRA scores of forgotten candidates in
$\mathcal{A}_{\mathrm U}$ from those of never-seen candidates in
$\mathcal{A}_{\mathrm N}$. Because the direction of the LiRA score may be
reversed after an unlearning update, a conventional AUC below $0.5$ does not
necessarily imply that the two groups are indistinguishable. We therefore
report the two-sided U/N AUC:
\begin{equation}
\operatorname{AUC}_{\mathrm{U/N}}^{\mathrm{two}}
=
\max
\left\{
a_{\mathrm{U/N}},
1-a_{\mathrm{U/N}}
\right\}.
\label{eq:two-sided-un-auc}
\end{equation}
The resulting metric lies in $[0.5,1]$. A value close to $0.5$ indicates that
the forgotten and never-seen candidates are difficult to distinguish, whereas
a larger value indicates that they remain separable regardless of the score
direction.

\textbf{RNMSE} Forecasting performance is evaluated after restoring both predictions and
targets to their original data scale. Let
$\mathcal{Q}=\{(\vx_t,\vy_t)\}_{t\in\sT_{\mathcal Q}}$ denote an evaluation
set, where $\sT_{\mathcal Q}$ is its set of forecast origins. For a forecasting
model $f_{\vtheta}$, its root mean squared error on $\mathcal Q$ is defined as
\begin{equation}
\operatorname{RMSE}_{\mathcal Q}
\left(
f_{\vtheta}
\right)
=
\sqrt{
\frac{
1
}{
\left|
\sT_{\mathcal Q}
\right|
hp
}
\sum_{t\in\sT_{\mathcal Q}}
\left\|
f_{\vtheta}
\left(
\vx_t
\right)
-
\vy_t
\right\|_{\mathrm F}^{2}
}.
\label{eq:forecast-rmse}
\end{equation}
Here, $h$ denotes the forecasting horizon and $p$ denotes the number of
variables. For the univariate setting considered in our notation, $p=1$; for
multivariate forecasting, the prediction and target of each window contain
$hp$ scalar values.

Let $\sigma_{\mathrm{train}}$ denote the standard deviation of the raw training
values used to train the original model, computed before applying the deletion
request. We normalize the forecasting error as
\begin{equation}
\operatorname{NRMSE}_{\mathcal Q}
\left(
f_{\vtheta}
\right)
=
\frac{
\operatorname{RMSE}_{\mathcal Q}
\left(
f_{\vtheta}
\right)
}{
\sigma_{\mathrm{train}}
}.
\label{eq:forecast-nrmse}
\end{equation}
The test NRMSE is obtained by setting $\mathcal Q=\train_{\mathrm{test}}$.
A lower NRMSE indicates better forecasting performance relative to the
variation of the original training series.

\textbf{F-P Gap} To measure the functional agreement between an evaluated model \(f\) and exact retraining on the windows affected by the deletion request, we define the Forgetting-set Prediction Gap as
\begin{equation}
\operatorname{F\text{-}P\,Gap}(f)
=
\frac{100 \%}{\sigma_{\mathrm{train}}}
\sqrt{
\frac{1}{|\train_{\mathrm f}|hp}
\sum_{(\vx_t,\vy_t)\in\train_{\mathrm f}}
\left\|
f(\vx_t)
-
f_{\vtheta_{\train_{\mathrm r}}^{*}}(\vx_t)
\right\|_{\mathrm F}^{2}
}.
\label{eq:fp-gap}
\end{equation}
Here, \(f_{\vtheta_{\train_{\mathrm r}}^{*}}\) is the model trained from scratch on the retained set \(\train_{\mathrm r}\), \(\sigma_{\mathrm{train}}\) is the standard deviation of the raw training targets before deletion, \(h\) is the forecasting horizon, and \(p\) is the number of variables. The factor \(100\) expresses the normalized gap as a percentage. Unlike forecasting NRMSE, which compares predictions with the ground-truth, the F-P Gap compares the predictions of \(f\) with those of exact retraining on the forgetting set \(\train_{\mathrm f}\). A lower F-P Gap indicates closer functional agreement with exact retraining, and a value of zero indicates identical predictions on all forgetting inputs.
\subsection{Baselines}
We compare RDTU with the following representative unlearning baselines. \textbf{Original} denotes the forecasting model before unlearning, while \textbf{Retrain} trains the same architecture from scratch on the strictly retained set and serves as the exact-unlearning reference. \textbf{SSD}~\citep{foster2024fast} uses Fisher information to identify parameters disproportionately associated with the forgetting set and selectively dampens them. \textbf{NegGrad+}~\citep{NEURIPS2023_062d711f} combines gradient ascent on the forgetting set with gradient descent on the retained set. \textbf{Zero-label} replaces the targets of affected windows with zero vectors, whereas \textbf{Random-label} assigns randomly generated continuous targets following the random-labeling strategy. \textbf{SCRUB-R}~\citep{NEURIPS2023_062d711f} adopts a teacher--student framework that moves the student away from the original model on the forgetting set while preserving its predictions on the retained set. \textbf{TS-Unlearn}~\citep{wang2026ts} constructs replacement targets by feeding Gaussian noise into the original model and dynamically balances the forgetting and retention objectives through Pareto optimization.

\section{More Experiments}
To demonstrate the robustness and applicability of our model, we conducted additional expansion experiments with different forecasting horizon lengths. Table~\ref{tab:robustness_h3_univariate} and Table~\ref{tab:robustness_h3_multivariate} respectively show the experimental results for horizon lengths of 3.

\begin{table*}[h]
\centering
\caption{
Robustness comparison on univariate datasets under a forecasting
horizon of $H=3$. U/N values closer to $0.5$ indicate better
unlearning effectiveness, while lower F-P Gap, NRMSE, and runtime
indicate better performance.
}
\label{tab:robustness_h3_univariate}

\setlength{\tabcolsep}{4.5pt}
\renewcommand{\arraystretch}{1.05}

\resizebox{\textwidth}{!}{%
\begin{tabular}{l cccc cccc}
\toprule
\textbf{Method}
& \multicolumn{4}{c}{\textbf{Air Quality}}
& \multicolumn{4}{c}{\textbf{Traffic}} \\
\cmidrule(lr){2-5}
\cmidrule(lr){6-9}

& U/N & F-P Gap & NRMSE & Time
& U/N & F-P Gap & NRMSE & Time \\
\midrule

Original
& 0.6082 & 10.91 & 0.4084 & --
& 0.6716 & 15.57 & 0.4908 & -- \\

Retrain
& 0.5400 & -- & 0.4131 & 20.480
& 0.5478 & -- & 0.5063 & 12.640 \\

\midrule

SSD
& 0.5429 & 43.31 & 0.6725 & 0.77
& 0.5501 & 79.35 & 0.9927 & 0.450 \\

NegGrad+
& 0.5356 & 167.58 & 1.6929 &7.88
& 0.5346 & 79.90 & 0.8708 & 3.930 \\

Zero-label
& 0.5348 & 50.79 & 0.6918 & 3.33
& 0.5507 & 49.89 & 0.7294 & 1.820 \\

Random-label
& 0.5367 & 50.51 & 0.6933 & 3.74
& 0.5490 & 50.05 & 0.7425 & 2.170 \\

SCRUB-R
& 0.5439 & 10.21 & 0.4077 & 8.43
& 0.5627 & 14.87 & 0.4995 & 4.400 \\

TS-Unlearn
& 0.5415 & 10.74 & 0.4128 & 6.43
& 0.5464 & 17.64 & 0.5077 & 3.780 \\

\midrule

\textbf{Ours}
& 0.5483 & 8.01 & 0.3977 & 10.95
& 0.5377 & 14.75 & 0.4970 & 4.550 \\

\bottomrule
\end{tabular}%
}
\end{table*}

\begin{table*}[h]
\centering
\caption{
Robustness comparison on multivariate datasets under a forecasting
horizon of $H=3$. U/N values closer to $0.5$ indicate better
unlearning effectiveness, while lower F-P Gap, NRMSE, and runtime
indicate better performance.
}
\label{tab:robustness_h3_multivariate}

\setlength{\tabcolsep}{4.5pt}
\renewcommand{\arraystretch}{1.05}

\resizebox{\textwidth}{!}{%
\begin{tabular}{l cccc cccc}
\toprule
\textbf{Method}
& \multicolumn{4}{c}{\textbf{EEG}}
& \multicolumn{4}{c}{\textbf{HAR70}} \\
\cmidrule(lr){2-5}
\cmidrule(lr){6-9}

& U/N & F-P Gap & NRMSE & Time
& U/N & F-P Gap & NRMSE & Time \\
\midrule

Original
& 0.6759 & 5.24 & 0.1986 & --
& 0.6946 & 12.53 & 0.3977 & -- \\

Retrain
& 0.5152 & -- & 0.2055 & 8.980
& 0.5162 & -- & 0.3975 & 30.71 \\

\midrule

SSD
& 0.5257 & 59.03 & 0.6593 & 0.230
& 0.5247 & 82.05 & 0.6331 & 0.55 \\

NegGrad+
& 0.5267 & 95.88 & 0.8748 & 2.350
& 0.5314 & 208.95 & 1.1839 & 9.28 \\

Zero-label
& 0.5244 & 60.22 & 0.6029 & 0.360
& 0.5231 & 58.71 & 0.4528 & 0.62 \\

Random-label
& 0.5247 & 60.02 & 0.6012 & 0.660
& 0.5235 & 45.24 & 0.5611 & 1.00 \\

SCRUB-R
& 0.5452 & 8.51 & 0.2191 & 2.810
& 0.5613 & 12.23 & 0.3976 & 9.92 \\

TS-Unlearn
& 0.5324 & 8.31 & 0.2155 & 0.930
& 0.5477 & 12.75 & 0.4016 & 1.49 \\

\midrule

\textbf{Ours}
& 0.5207 & 6.68 & 0.2107 & 2.670
& 0.5305 & 10.61 & 0.3968 & 7.61 \\

\bottomrule
\end{tabular}%
}
\end{table*}
\textbf{\textit{The results show that RDTU remains effective when the forecasting horizon is
increased to 3. In particular, RDTU achieves the lowest F-P Gap among all
approximate-unlearning methods on all four datasets.}} Compared with the strongest
baseline for each dataset, RDTU reduces the F-P Gap by $21.5\%$, $0.8\%$,
$19.6\%$, and $13.2\%$ on Air Quality, Traffic, EEG, and HAR70, respectively.
The improvement is particularly pronounced on Air Quality and EEG, indicating
that the counterfactual labels generated by RDTU remain close to the predictions
of exact retraining even when each prediction contains multiple future steps.

\textbf{\textit{RDTU also obtains the lowest test NRMSE among the approximate-unlearning methods
across all datasets,}} reducing NRMSE by approximately $0.2\%$--$2.5\%$ relative
to the best-performing baseline. Its U/N AUC values remain between $0.5207$ and
$0.5483$, demonstrating competitive membership indistinguishability without the
severe forecasting degradation observed for SSD, NegGrad+, Zero-label, and
Random-label. For example, some of these methods obtain U/N AUC values close to $0.5$, but simultaneously produce F-P Gaps above $40\%$ and substantially larger NRMSE values. \textbf{\textit{This suggests that their apparent membership indistinguishability is partly caused by excessive disruption of the forecasting function rather than close approximation to exact retraining.}}

We also conducted experiments with a horizon length of 6. The results of these experiments are presented in Tables~\ref{tab:robustness_h6_univariate} and Tables~\ref{tab:robustness_h6_multivariate}. The results show that in this configuration, our model also achieved excellent results.

\begin{table*}[h]
\centering
\caption{
Robustness comparison on univariate datasets under a forecasting
horizon of $H=6$. U/N values closer to $0.5$ indicate better
unlearning effectiveness, while lower F-P Gap, NRMSE, and runtime
indicate better performance.
}
\label{tab:robustness_h6_univariate}

\setlength{\tabcolsep}{4.5pt}
\renewcommand{\arraystretch}{1.05}

\resizebox{\textwidth}{!}{%
\begin{tabular}{l cccc cccc}
\toprule
\textbf{Method}
& \multicolumn{4}{c}{\textbf{Air Quality}}
& \multicolumn{4}{c}{\textbf{Traffic}} \\
\cmidrule(lr){2-5}
\cmidrule(lr){6-9}

& U/N & F-P Gap & NRMSE & Time
& U/N & F-P Gap & NRMSE & Time \\
\midrule

Original
& 0.6653 & 12.39 & 0.5379 & --
& 0.6645 & 17.02 & 0.5821 & --\\

Retrain
& 0.5330 & -- & 0.5556 & 16.54
& 0.5698 & -- & 0.5916 & 10.94\\

\midrule

SSD
& 0.5312 & 51.25 & 0.7903 & 1.02
& 0.5661 & 78.50 & 0.9965 & 0.76 \\

NegGrad+
& 0.5423 & 105.31 & 1.1596 & 7.48
& 0.5583 & 53.88 & 0.7481 & 3.53 \\

Zero-label
& 0.5310 & 50.57 & 0.7553 & 4.14
& 0.5723 & 49.50 & 0.7698 & 2.84 \\

Random-label
& 0.5304 & 49.75 & 0.7516 & 3.92
& 0.5745 & 49.46 & 0.7805 & 2.47 \\

SCRUB-R
& 0.5360 & 46.37 & 0.6910 & 5.98
& 0.5732 & 51.38 & 0.8780 & 3.16 \\

TS-Unlearn
& 0.5562 & 11.69 & 0.5494 & 8.01
& 0.5743 & 30.89 & 0.6613 & 4.99 \\

\midrule

\textbf{Ours}
& 0.5521 & 9.95 & 0.5450 & 12.44 
& 0.5698 & 16.09 & 0.5803 & 7.54\\

\bottomrule
\end{tabular}%
}
\end{table*}

\begin{table*}[h]
\centering
\caption{
Robustness comparison on multivariate datasets under a forecasting
horizon of $H=6$. U/N values closer to $0.5$ indicate better
unlearning effectiveness, while lower F-P Gap, NRMSE, and runtime
indicate better performance.
}
\label{tab:robustness_h6_multivariate}

\setlength{\tabcolsep}{4.5pt}
\renewcommand{\arraystretch}{1.05}

\resizebox{\textwidth}{!}{%
\begin{tabular}{l cccc cccc}
\toprule
\textbf{Method}
& \multicolumn{4}{c}{\textbf{EEG}}
& \multicolumn{4}{c}{\textbf{HAR70}} \\
\cmidrule(lr){2-5}
\cmidrule(lr){6-9}

& U/N & F-P Gap & NRMSE & Time
& U/N & F-P Gap & NRMSE & Time \\
\midrule

Original
& 0.7301 & 10.74 & 0.3678 & --
& 0.6982 & 13.88 & 0.4393 & -- \\

Retrain
& 0.5210 & -- & 0.3770 & --
& 0.5215 & -- & 0.4430 & 34.45 \\

\midrule

SSD
& 0.5268 & 58.46 & 0.7230 & 0.62
& 0.5187 & 81.82 & 0.6467 & 0.86 \\

NegGrad+
& 0.5305 & 93.27 & 0.9043 & 2.60
& 0.5271 & 182.69 & 0.9548 & 9.91 \\

Zero-label
& 0.5289 & 56.45 & 0.6505 & 0.79
& 0.5199 & 59.14 & 0.4808 & 1.10 \\

Random-label
& 0.5284 & 56.31 & 0.6489 & 0.36
& 0.5241 & 45.48 & 0.5931 & 0.67 \\

SCRUB-R
& 0.5653 & 13.49 & 0.3872 & 2.83
& 0.5596 & 13.50 & 0.4407 & 4.93 \\

TS-Unlearn
& 0.5458 & 14.57 & 0.3883 & 1.20
& 0.5460 & 14.57 & 0.4431 & 1.78 \\

\midrule

\textbf{Ours}
& 0.5348 & 11.02 & 0.3792 & 12.97
& 0.5361 & 11.26 & 0.4379 & 8.23 \\

\bottomrule
\end{tabular}%
}
\end{table*}

\section{More Hyperparameter Analysis}
\subsection{Influence of Diffusion Steps on Different Horizon Length}
\begin{figure}[h]
    \centering
    \includegraphics[width=\linewidth]{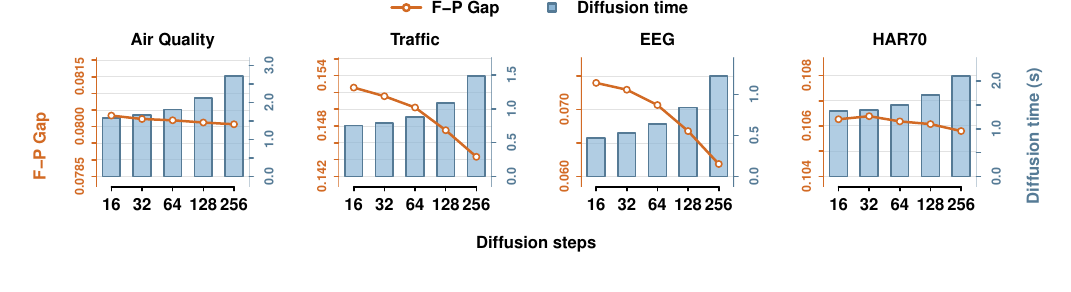}
    \vspace{-5mm}
    \caption{
    Sensitivity analysis of RDTU under different diffusion steps
   of Horizon Length=3.
    The curves report the F-P Gap and the time cost of diffusion.
    }
    \label{fig:step_3}
\end{figure}

The same with the setting where the horizon length is 1, we also examined the impact of the diffusion step number on the model with a horizon length of 3. Figure~\ref{fig:step_3} presents the detailed experimental results. \textbf{\textit{From the figure, we can observe that the trend of the influence of the diffusion step count on the model's performance and training duration is consistent with the case where the horizon is 1.}} Therefore, we have set the diffusion step count here to 128 as well.

\subsection{Influence of Forget Loss Weight}
\begin{figure}[h]
    \centering
    \includegraphics[width=\linewidth]{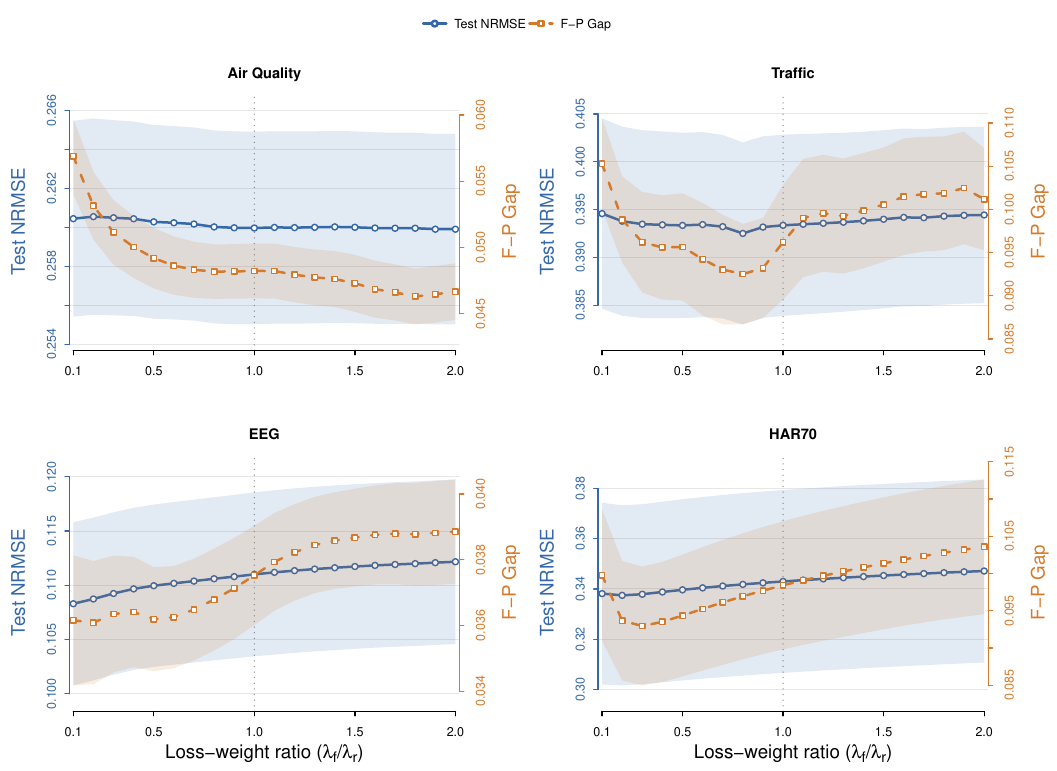}
    \caption{
    Sensitivity analysis of RDTU under different loss-weight ratios
    $\lambda_{\mathrm f}/\lambda_{\mathrm r}$ of Horizon Length=1.
    The curves report the mean test NRMSE and F-P Gap, while the shaded
    regions indicate the corresponding standard errors over 30 runs.
    Lower values indicate better performance for both metrics.
    }
    \label{fig:loss-weight-ratio_1}
\end{figure}

Figure~\ref{fig:loss-weight-ratio_1} examines the effect of the loss-weight ratio
$\lambda_{\mathrm f}/\lambda_{\mathrm r}$. We select the ratio primarily according to the F-P Gap and NRMSE, because it directly measures the functional agreement between the unlearned model and the exact-retrained model on the affected
windows and retain set. A lower F-P Gap therefore indicates that the generated labels guide
the model more closely toward the desired post-deletion behavior, and a lower RNMSE represents the model maintained good performance. Based on this criterion, the best ratios are $1.8$, $0.8$, $0.2$, and $0.3$ for Air Quality, Traffic, EEG, and HAR70, respectively. Overall, the corresponding test NRMSE changes remain relatively moderate, indicating that optimizing the F-P Gap does not cause substantial degradation in forecasting utility. \textbf{\textit{These results also show that the preferred balance between the forgetting loss and the retained anchoring loss varies across datasets because of their different temporal structures and prediction scales.}}

\section{Future Work}
\subsection{Computational scalability}
Although RDTU is more efficient than exact retraining, its computational cost may increase for long sequences, high-dimensional multivariate data, and large forecasting architectures. In particular, empirical NTK construction requires computing and storing parameter-gradient features for numerous forecasting windows, while residual generation involves multiple denoising steps for every affected query. Future work could reduce these costs through incremental Nystr"om approximation, low-rank kernel representations, and cached NTK features that can be reused across deletion requests. The retained-reference statistics required by SIS could likewise be updated incrementally instead of being recomputed after each deletion. On the diffusion side, accelerated or distilled samplers, batched residual generation, and adaptive denoising schedules could further reduce inference time. For example, queries with strong retained support could use fewer denoising steps, whereas structurally irreplaceable queries with high SIS values could receive additional computation. A comprehensive scalability study should separately report one-time preprocessing costs, per-request latency, peak memory consumption, and amortized runtime as the number of windows, variables, deletion requests, and diffusion steps increases.

\subsection{Broader evaluation and stronger guarantees}
Future evaluations should examine RDTU under a broader range of forecasting and deletion conditions. Relevant settings include longer forecasting horizons, higher-dimensional multivariate series, irregular sampling, missing observations, temporal distribution shifts, and deletion intervals spanning rare or rapidly changing patterns. It would also be valuable to vary the deletion ratio, interval length, and number of sequential requests to determine when the residual-transfer assumption begins to deteriorate. Evaluations with recurrent networks, temporal convolutional networks, Transformers, and state-space models would further assess the generality of the architecture-agnostic design. In addition, the current evaluation measures empirical agreement with exact retraining and resistance to membership inference, but these results do not provide a formal guarantee of data removal. Future studies should consider adaptive black-box and white-box attacks that exploit model outputs, gradients, parameters, or intermediate representations, together with reconstruction and model-inversion attacks. A complementary theoretical direction is to bound the discrepancy between the unlearned and exact-retrained models in terms of the NTK approximation error, residual-generation error, and optimization error. Such analyses could support verifiable stopping criteria or certified variants of RDTU for deployment in privacy-sensitive applications.

%% file: iclr2027_conference.bib
@article{zhao2025exploring,
  title={Exploring federated unlearning: Review, comparison, and insights},
  author={Zhao, Yang and Yang, Jiaxi and Tao, Yiling and Wang, Lixu and Li, Xiaoxiao and Niyato, Dusit and Poor, H Vincent},
  journal={IEEE Network},
  volume={40},
  number={2},
  pages={175--183},
  year={2025},
  publisher={IEEE}
}

@inproceedings{lai2018modeling,
  title     = {Modeling Long- and Short-Term Temporal Patterns with Deep Neural Networks},
  author    = {Lai, Guokun and Chang, Wei-Cheng and Yang, Yiming and Liu, Hanxiao},
  booktitle = {Proceedings of the 41st International ACM SIGIR Conference on Research and Development in Information Retrieval},
  year      = {2018}
}

@article{zyma2019electroencephalograms,
  title   = {Electroencephalograms during Mental Arithmetic Task Performance},
  author  = {Zyma, Igor and Tukaev, Sergii and Seleznov, Ivan
             and Kiyono, Ken and Popov, Anton
             and Chernykh, Mariia and Shpenkov, Oleksii},
  journal = {Data},
  volume  = {4},
  number  = {1},
  pages   = {14},
  year    = {2019},
  doi     = {10.3390/data4010014}
}

@misc{har70plus_780,
  author       = {Logacjov, Aleksej and Ustad, Astrid},
  title        = {{HAR70+}},
  year         = {2023},
  howpublished = {UCI Machine Learning Repository},
  doi          = {10.24432/C5CW3D},
}

@inproceedings{NEURIPS2021_bcc0d400,
 author = {Wu, Haixu and Xu, Jiehui and Wang, Jianmin and Long, Mingsheng},
 booktitle = {Advances in Neural Information Processing Systems},
 title = {Autoformer: Decomposition Transformers with Auto-Correlation for Long-Term Series Forecasting},
 year = {2021}
}

@InProceedings{pmlr-v119-guo20c,
  title = 	 {Certified Data Removal from Machine Learning Models},
  author =       {Guo, Chuan and Goldstein, Tom and Hannun, Awni and Van Der Maaten, Laurens},
  booktitle = 	 {Proceedings of the 37th International Conference on Machine Learning},
  year = 	 {2020},
}

@inproceedings{NEURIPS2019_cb79f8fa,
 author = {Ginart, Antonio and Guan, Melody and Valiant, Gregory and Zou, James},
 booktitle = {Advances in Neural Information Processing Systems},
 title = {Making AI Forget You: Data Deletion in Machine Learning},
 year = {2019}
}

@inproceedings{
zhang2024negative,
title={Negative Preference Optimization: From Catastrophic Collapse to Effective Unlearning},
author={Ruiqi Zhang and Licong Lin and Yu Bai and Song Mei},
booktitle={First Conference on Language Modeling},
year={2024}
}

@inproceedings{graves2021amnesiac,
  title={Amnesiac machine learning},
  author={Graves, Laura and Nagisetty, Vineel and Ganesh, Vijay},
  booktitle={Proceedings of the AAAI conference on artificial intelligence},
  year={2021}
}

@inproceedings{NEURIPS2018_5a4be1fa,
 author = {Jacot, Arthur and Gabriel, Franck and Hongler, Clement},
 booktitle = {Advances in Neural Information Processing Systems},
 title = {Neural Tangent Kernel: Convergence and Generalization in Neural Networks},
 year = {2018}
}

@inproceedings{NEURIPS2019_0d1a9651,
 author = {Lee, Jaehoon and Xiao, Lechao and Schoenholz, Samuel and Bahri, Yasaman and Novak, Roman and Sohl-Dickstein, Jascha and Pennington, Jeffrey},
 booktitle = {Advances in Neural Information Processing Systems},
 title = {Wide Neural Networks of Any Depth Evolve as Linear Models Under Gradient Descent},
 year = {2019}
}

@inproceedings{NEURIPS2020_ad086f59,
 author = {Lee, Jaehoon and Schoenholz, Samuel and Pennington, Jeffrey and Adlam, Ben and Xiao, Lechao and Novak, Roman and Sohl-Dickstein, Jascha},
 booktitle = {Advances in Neural Information Processing Systems},
 title = {Finite Versus Infinite Neural Networks: an Empirical Study},
 year = {2020}
}

@inproceedings{chundawat2023can,
  title={Can bad teaching induce forgetting? unlearning in deep networks using an incompetent teacher},
  author={Chundawat, Vikram S and Tarun, Ayush K and Mandal, Murari and Kankanhalli, Mohan},
  booktitle={Proceedings of the AAAI Conference on Artificial Intelligence},
  year={2023}
}

@article{williams2000using,
  title={Using the Nystr{\"o}m method to speed up kernel machines},
  author={Williams, Christopher and Seeger, Matthias},
  journal={Advances in neural information processing systems},
  volume={13},
  year={2000}
}

@article{lakshminarayanan2017simple,
  title={Simple and scalable predictive uncertainty estimation using deep ensembles},
  author={Lakshminarayanan, Balaji and Pritzel, Alexander and Blundell, Charles},
  journal={Advances in neural information processing systems},
  volume={30},
  year={2017}
}

@inproceedings{NEURIPS2023_062d711f,
 author = {Kurmanji, Meghdad and Triantafillou, Peter and Hayes, Jamie and Triantafillou, Eleni},
 booktitle = {Advances in Neural Information Processing Systems},
 title = {Towards Unbounded Machine Unlearning},
 year = {2023}
}

@article{10.1098/rsta.2020.0209,
    author = {Lim, Bryan and Zohren, Stefan},
    title = {Time-series forecasting with deep learning: a survey},
    journal = {Philosophical Transactions of the Royal Society A: Mathematical, Physical and Engineering Sciences},
    volume = {379},
    pages = {20200209},
    year = {2021},
    doi = {10.1098/rsta.2020.0209},
}

@article{zonta2020predictive,
  title={Predictive maintenance in the Industry 4.0: A systematic literature review},
  author={Zonta, Tiago and Da Costa, Cristiano Andr{\'e} and da Rosa Righi, Rodrigo and De Lima, Miromar Jos{\'e} and Da Trindade, Eduardo Silveira and Li, Guann Pyng},
  journal={Computers \& industrial engineering},
  volume={150},
  pages={106889},
  year={2020}
}

@article{eu2016general,
  title={General data protection regulation},
  author={EU, GDPR},
  journal={Official Journal of the European Union},
  year={2016}
}

@article{teh2020sensor,
  title={Sensor data quality: A systematic review},
  author={Teh, Hui Yie and Kempa-Liehr, Andreas W and Wang, Kevin I-Kai},
  journal={Journal of Big Data},
  volume={7},
  pages={11},
  year={2020}
}

@article{goknil2023systematic,
  title={A systematic review of data quality in CPS and IoT for industry 4.0},
  author={Goknil, Arda and Nguyen, Phu and Sen, Sagar and Politaki, Dimitra and Niavis, Harris and Pedersen, Karl John and Suyuthi, Abdillah and Anand, Abhilash and Ziegenbein, Amina},
  journal={ACM Computing Surveys},
  volume={55},
  pages={1--38},
  year={2023}
}

@inproceedings{cao2015towards,
  title={Towards making systems forget with machine unlearning},
  author={Cao, Yinzhi and Yang, Junfeng},
  booktitle={2015 IEEE symposium on security and privacy},
  year={2015},
}

@inproceedings{
maini2024tofu,
title={{TOFU}: A Task of Fictitious Unlearning for {LLM}s},
author={Pratyush Maini and Zhili Feng and Avi Schwarzschild and Zachary Chase Lipton and J Zico Kolter},
booktitle={First Conference on Language Modeling},
year={2024},
}

@inproceedings{wang2026ts,
  title={TS-Unlearn: A Dual-Objective Unlearning Framework for Time Series Forecasting},
  author={Wang, Ziyi and Chen, Lixing and Miao, Zhongqi and Tang, Junhua and Bai, Yang and Li, Jianhua},
  booktitle={Pacific-Asia Conference on Knowledge Discovery and Data Mining},
  year={2026},
}

@inproceedings{foster2024fast,
  title={Fast machine unlearning without retraining through selective synaptic dampening},
  author={Foster, Jack and Schoepf, Stefan and Brintrup, Alexandra},
  booktitle={Proceedings of the AAAI conference on artificial intelligence},
  year={2024}
}

@inproceedings{di2026label,
  title={Label smoothing improves machine unlearning},
  author={Di, Zonglin and Zhu, Zhaowei and Jia, Jinghan and Liu, Jiancheng and Takhirov, Zafar and Jiang, Bo and Yao, Yuanshun and Liu, Sijia and Liu, Yang},
  booktitle={International Conference on Learning Representations},
  year={2026}
}

@article{chizat2019lazy,
  title={On lazy training in differentiable programming},
  author={Chizat, Lenaic and Oyallon, Edouard and Bach, Francis},
  journal={Advances in neural information processing systems},
  volume={32},
  year={2019}
}

@inproceedings{yang2021tensor,
  title={Tensor programs iv: Feature learning in infinite-width neural networks},
  author={Yang, Greg and Hu, Edward J},
  booktitle={International Conference on Machine Learning},
  year={2021},
}

@inproceedings{shokri2017membership,
  title={Membership inference attacks against machine learning models},
  author={Shokri, Reza and Stronati, Marco and Song, Congzheng and Shmatikov, Vitaly},
  booktitle={2017 IEEE symposium on security and privacy (SP)},
  year={2017},
}

@inproceedings{carlini2022membership,
  title={Membership inference attacks from first principles},
  author={Carlini, Nicholas and Chien, Steve and Nasr, Milad and Song, Shuang and Terzis, Andreas and Tramer, Florian},
  booktitle={2022 IEEE symposium on security and privacy (SP)},
  year={2022},
}

@article{vaswani2017attention,
  title={Attention is all you need},
  author={Vaswani, Ashish and Shazeer, Noam and Parmar, Niki and Uszkoreit, Jakob and Jones, Llion and Gomez, Aidan N and Kaiser, {\L}ukasz and Polosukhin, Illia},
  journal={Advances in neural information processing systems},
  volume={30},
  year={2017}
}

@misc{beijing_multi-site_air_quality_501,
  author       = {Chen, Song},
  title        = {{Beijing Multi-Site Air Quality}},
  year         = {2017},
  howpublished = {UCI Machine Learning Repository},
  DOI = {https://doi.org/10.24432/C5RK5G}
}

@article{Mohapatra2024PhasedrivenGR,
  title={Phase-driven Generalizable Representation Learning for Nonstationary Time Series Classification},
  author={Payal Mohapatra and Lixu Wang and Qi Zhu},
  journal={Trans. Mach. Learn. Res.},
  year={2024},
  volume={2025},
}

@inproceedings{
gao2025on,
title={On Large Language Model Continual Unlearning},
author={Chongyang Gao and Lixu Wang and Kaize Ding and Chenkai Weng and Xiao Wang and Qi Zhu},
booktitle={The Thirteenth International Conference on Learning Representations},
year={2025},
}

@inproceedings{wang2024dacr,
  title={Dacr: distribution-augmented contrastive reconstruction for time-series anomaly detection},
  author={Wang, Lixu and Xu, Shichao and Du, Xinyu and Zhu, Qi},
  booktitle={ICASSP 2024-2024 IEEE International Conference on Acoustics, Speech and Signal Processing (ICASSP)},
  pages={7545--7549},
  year={2024},
}

@inproceedings{
wang2022nontransferable,
title={Non-Transferable Learning: A New Approach for Model Ownership Verification and Applicability Authorization},
author={Lixu Wang and Shichao Xu and Ruiqi Xu and Xiao Wang and Qi Zhu},
booktitle={International Conference on Learning Representations},
year={2022},
}
